\documentclass[review]{elsarticle}
\usepackage{lineno,hyperref}
\usepackage{multirow}
\usepackage{multicol}
\usepackage{xcolor}
\usepackage{tabularx}
\usepackage{graphicx}
\usepackage{ragged2e}
\usepackage{amsmath}
\usepackage{amsfonts}
\usepackage{amssymb}

\modulolinenumbers[5]
\usepackage[ruled,vlined,linesnumbered,noresetcount]{algorithm2e}
\usepackage{algpseudocode}
\usepackage{gensymb}
\journal{Journal of \LaTeX\ Templates}
\usepackage{textcomp}
\usepackage[superscript]{cite}

\begin{document}

% ---------------- TITLE PAGE START ----------------
\title{Physics-Informed Neural Network Approaches for Sparse Data Flow Reconstruction of Unsteady Flow Around Complex Geometries }

%% Authors and affiliations
\author[mymainaddress]{Vamsi Sai Krishna Malineni}
\author[mymainaddress]{Suresh Rajendran}
\address[mymainaddress]{Department of Ocean Engineering, Indian Institute of Technology, Madras, \\  Chennai - 600036. INDIA}
% ---------------- TITLE PAGE END ----------------

\begin{frontmatter}

% ---------------- ABSTRACT START ----------------
\begin{abstract}
The utilization of Deep Neural Networks (DNNs) in physical science and engineering applications has gained traction due to their capacity to learn intricate functions. While large datasets are crucial for training DNN models in fields like computer vision and natural language processing, obtaining such datasets for engineering applications is prohibitively expensive. Physics-Informed Neural Networks (PINNs), a branch of Physics-Informed Machine Learning (PIML), tackle this challenge by embedding physical principles within neural network architectures. PINNs have been extensively explored for solving diverse forward and inverse problems in fluid mechanics. Nonetheless, there is limited research on employing PINNs for flow reconstruction from sparse data under constrained computational resources. Earlier studies were focused on forward problems with well-defined data. The present study attempts to develop models capable of reconstructing the flow field data from sparse datasets mirroring real-world scenarios.

This study focuses on two cases: (a) two-dimensional (2D) unsteady laminar flow past a circular cylinder and (b) three-dimensional (3D) unsteady turbulent flow past an ultra-large container ship (ULCS). The first case compares the effectiveness of training methods like Standard PINN and Backward Compatible PINN (BC-PINN) and explores the performance enhancements through systematic relaxation of physics constraints and dynamic weighting of loss function components. The second case highlights the capability of PINN-based models to learn underlying physics from sparse data while accurately reconstructing the flow field for a highly turbulent flow.
\end{abstract}
% ---------------- ABSTRACT END ----------------


\end{frontmatter}

%%%%%%%%%%%%%%%%%%%%%%%%%%%%%%%%%%%%%%%%%%%%%%%%%%%%%%%%%%%%%%%%%%%%%
%%%%%%%%%%%%%%%% Summary of Notation %%%%%%%%%%%%%%%%%%%%%%%%%%%%%%%%
%%%%%%%%%%%%%%%%%%%%%%%%%%%%%%%%%%%%%%%%%%%%%%%%%%%%%%%%%%%%%%%%%%%%%
\newpage 
\begin{center}
    {\textbf{\large{Summary of Notations}}}
\end{center}
\newcolumntype{C}[1]{>{\raggedright\arraybackslash}p{#1}}
\begin{tabular}{C{1.2cm}C{4cm}C{1.2cm}C{4cm}}
 $PDE$ & Partial Differential Equation  & $MSE$ & Mean Square Error\\
 $FC-FFNN$ & Fully Connected Feed-Forward Neural Network  & $PIML$ & Physics-informed Machine Learning\\
 $PINN$ & Physics informed neural network &$ RANS $& Reynolds Averaged Navier Stokes Equations \\
 $CFD$ & Computational Fluid Dynamics &$ SGD $&Stochastic Gradient Descent \\
 $AD$& Automatic Differentiation & $BC-PINN$& Backward Compatible Physics Informed Neural Network \\

\end{tabular}
%%%%%%%%%%%%%%%%%%%%%%%%%%%%%%%%%%%%%%%%%%%%%%%%%%%%%%%%%%%

%%%%%%%%%%%%%%%%%%%%%%%%%%%%%%%%%%%%%%%%%%%%%%%%%%%%%%%%%%%%%%%%%%%%%
%%%%%%%%%%%%%%%%%%%%%%% Introduction %%%%%%%%%%%%%%%%%%%%%%%%%%%%%%%%
%%%%%%%%%%%%%%%%%%%%%%%%%%%%%%%%%%%%%%%%%%%%%%%%%%%%%%%%%%%%%%%%%%%%%
\newpage
\section*{I. INTRODUCTION}
\addcontentsline{toc}{section}{I. INTRODUCTION}
\justifying
Recent breakthroughs in machine learning (ML) have revitalized the enthusiasm of the fluid mechanics community for the development of computationally efficient surrogate models. These models are aimed at addressing complex, ill-posed inverse problems while also being capable of real-time data retrieval applications\cite{vinuesa2022emerging},\cite{vinuesa2022enhancing}\cite{brunton2020machine}. ML-based approaches can be broadly classified into two: (i) data-driven approach \cite{vinuesa2022enhancing} and (ii) physics-informed approach \cite{brunton2020machine},\cite{willard2022integrating},\cite{karniadakis2021physics},\cite{cuomo2022scientific}.Although data-driven methods in ML have seen notable progress, they typically demand extensive datasets to effectively capture the fundamental dynamics of complex physical systems.\\
\justifying 
Data-driven approaches operate without reliance on physical principles, aiming to infer the underlying physics directly from the data. In contrast, physics-informed approaches incorporate partial or complete knowledge of the problem's underlying physics into the neural network framework. This study emphasizes the physics-informed methodology, particularly Physics-Informed Neural Networks (PINNs) \cite{raissi2019physics}, which represent a specialized category of deep neural networks [\cite{goodfellow2016deep}]. PINNs embed the governing partial differential equation (PDE) of the underlying physics through loss function \cite{raissi2019physics}. This physics loss component of the loss function acts as a regularizer to ensure that the model predictions are consistent with the physical laws. This also enables solving varied forward and inverse problems with minimal change in the network architecture \cite{raissi2019physics}.PINNs have been successfully applied to a variety of ill-posed inverse problems, including data-driven discovery of partial differential equations (PDEs) \cite{raissi2019physics}, integration of observational data through data assimilation \cite{raissi2019deep}, and the identification of underlying physical phenomena \cite{raissi2020hidden}. PINNs, once trained on a given spatiotemporal domain, allow for data retrieval from any spatiotemporal location within the given domain. The fact that only model parameters are required for data retrieval tasks after training makes PINNs highly memory efficient. PINNs are extremely flexible and can be implemented using well-developed deep learning \cite{goodfellow2016deep} software platforms that offer automatic differentiation feature \cite{baydin2018automatic} such as Tensorflow \cite{abadi2016tensorflow}, Pytorch \cite{paszke2019pytorch} and Jax \cite{bradbury2018jax}. \\

Standard PINNs, as proposed in \cite{raissi2019physics}, are difficult to train for forward problems. Standard PINNs have been analyzed in \cite{wang2021understanding} and \cite{krishnapriyan2021characterizing}, revealing two primary categories of failure modes. These are (i) Competing Optimization Objectives and (ii) Propagation failures. Rectification measures have been suggested to overcome these failure modes, such as adaptive weighting of the components of loss function [\cite{wang2021understanding},\cite{jin2021nsfnets}], using modified architectures \cite{wang2021understanding}, adaptive sampling of collocation points \cite{wu2023comprehensive} and sequential learning [\cite{krishnapriyan2021characterizing}]. Furthermore, \cite{gopakumar2023loss} demonstrates that PINNs are comparatively easier to train when dealing with sparse or coarse simulation outputs, as well as partially observed experimental data. In \cite{yang2023investigation}, PINN-based models were used for flow reconstruction from the data obtained from the Particle image velocimetry (PIV) method.
Several studies in the field of vortex-induced vibrations have also been reported (\cite{cheng2021deep},\cite{tang2022transfer},\cite{bai2022machine}). In \cite{cheng2021deep}, a PINN-based model was employed to solve the vortex-induced vibration (VIV) and the wake-induced vibration of a cylinder. In \cite{tang2022transfer}, a transfer-learning enhanced PINN-based model was proposed to study a 2D VIV system demonstrating time-saving capability. In \cite{bai2022machine}, a PINN is used for reconstructing VIV flows in turbulence regimes. In \cite{jagtap2022deep}, PINNs were used for solving ill-posed water wave problems governed by the Serre–Green–Naghdi equations by using multi-fidelity data for training the model. In \cite{eivazi2022physics}, PINNs were employed to solve the Reynolds Averaged Navier Stokes (RANS) equations for turbulent flows over a NACA4412 airfoil and the periodic hill. \\

This study considers two test cases: (a) 2D unsteady laminar flow past a circular cylinder and (b) 3D unsteady turbulent flow past a ship hull. The first case study is a benchmark problem. It has been extensively discussed in the past in the context of a forward problem, where PINN-based models are used to solve the governing PDE for a given spatiotemporal domain. However, past studies have not discussed flow reconstruction from sparse datasets. Moreover, the second case study shows the capability of the proposed methodology to predict complex turbulent flows like ship wake flow from sparse data set. The significant contributions of this article are:
\begin{itemize}
    \item %This study introduces a novel methodology to reconstruct flow fields from sparse datasets using Physics-Informed Neural Networks (PINNs).
    Unlike previous studies focused on forward problems with dense data, the proposed approach addresses the more challenging task of reconstructing complex flow fields, specifically for laminar and turbulent cases, from minimal data points, making it applicable to real-world engineering scenarios where data collection is limited. The method is able to predict the pressure accurately which is latent information in the flow field. 
    \item %By analyzing back-propagated gradients of the data and residual components within the loss function,
    This study identifies the limitations of standard PINNs for sparse data applications. It uses an adaptive weighting strategy that improves gradient flow, ensuring balanced learning from both sparse data and underlying physics. %This strategy addresses inherent issues in standard PINNs, where skewed gradients may bias the prediction, making it a significant contribution to the optimization of PINN models.
   
    %\item %Evaluation of BC-PINN vs. Standard PINN for Sparse Data: 
    %The study offers the first in-depth investigation into why Backward-Compatible PINN (BC-PINN) outperforms standard PINNs and data-driven models in scenarios with very sparse data and limited computational budgets. This work is the first to quantify and explain the performance benefits of BC-PINNs in sparse data applications.
    \item 3D Flow Reconstruction in Highly Turbulent Ship Wake**: The study extends the application of PINN-based models to reconstruct 3D turbulent wake flow behind an Ultra Large Container Ship (ULCS) from sparsely sampled data. Unlike laminar flow in simpler geometries, this case involves high-Reynolds-number turbulent flow, for which no previous studies have demonstrated the effective use of PINNs for such complex reconstructions, marking a pioneering effort in using PINNs for high-turbulence flow prediction around complex, real-world geometries.
    \item  %Resource-Efficient Flow Prediction:
    Recognizing computational constraints, this work demonstrates that PINN models, particularly the optimized BC-PINN, can accurately reconstruct flow fields with limited computational resources.% This resource-efficient approach broadens the applicability of PINNs to engineering problems where computational power is restricted, making advanced flow prediction methods accessible in more constrained environments.
\end{itemize}

The paper is organized as follows. The results of the 2D and 3D flow are discussed in sections 2 and 3, respectively.  Section 2 is subdivided into Dataset generation (2.1), Data-driven modeling (2.2), Physics informed modeling (2.3), and multi-objective optimization (2,4). General information such as CFD model setup (2.1.1), 	data set generation (2.1.2), and computation resources (2.1.3) are discussed in section 2.1. The particulars of the data-driven modeling (section 2.2), such as loss function formulation, hyperparameter tuning, and performance on the test data set, are mentioned in subsections 2.2.1, 2.2.2, and 2.2.3, respectively.   Similarly, the particulars of the physics-informed modeling (section 2.3), such as Loss function, hyperparameter tuning, and performance on the test data set, are mentioned in subsections 2.3.1, 2.3.2, and 2.3.3, respectively. The key aspects of the multi-objective optimization associated with the PINN formulations are discussed in section 2.4. Different Loss weighting strategies and their outcome is discussed under this section. 
Similarly, the CFD model set up, data set preparation, Loss function formulation and the PINN model training for the 3D wake flow problem is described in section 3.
 Finally, \ref{conclusion} summarizes the key findings and outlines potential directions for future research.

\section*{II. 2D UNSTEADY LAMINAR FLOW PAST A CYLINDER} \label{cylinder}
\addcontentsline{toc}{section}{II. 2D UNSTEADY LAMINAR FLOW PAST A CYLINDER}
In this section, we describe the first case study of a 2D unsteady laminar flow past a cylinder and the CFD simulations that are performed. Here, the wake prediction behind a cylinder is carried out using 3 methods, viz. 1) a purely data-driven model based on supervised learning, 2) standard PINN, and 3) BC-PINN. This section is organized as follows: Section \ref{2.1} primarily focuses on the setup of the computational fluid dynamics (CFD) model and the generation of labeled datasets. Section \ref{2.2} outlines the formulation of the loss function, details the process of hyper-parameter tuning, and primarily focuses on evaluating the performance of purely data-driven models using the test dataset. In section \ref{2.3}, physics-informed modeling is discussed, focusing on the formulation of the loss function, hyper-parameter tuning, and the performance evaluation of physics-informed surrogate models using the test dataset. Section \ref{2.4} elaborates on the case of competing objectives in the multiobjective optimization problem of training the physics-informed models for the current case study.

\subsection*{A. LABEL DATASET ACQUISITION AND DATASET GENERATION} \label{2.1}
\addcontentsline{toc}{subsection}{A. LABEL DATASET ACQUISITION AND DATASET GENERATION}
Acquiring large datasets for engineering applications is prohibitively expensive, given that the data comes from large-scale computational models or expensive experiments, as discussed in \cite{karniadakis2021physics}. It is an established fact that typical neural networks require large datasets to train and make proper predictions. However, the real-world dataset is characterized by sparsity owing to equipment limitations and monitoring technologies. To replicate real-world conditions, we employ a process of randomly sampling spatial points within the spatial domain. Data is gathered from these spatially distributed points at every time step. 

\subsubsection*{1. CFD MODEL SETUP}
\begin{figure*}[h!]
     \centering
     \includegraphics[width=\linewidth]{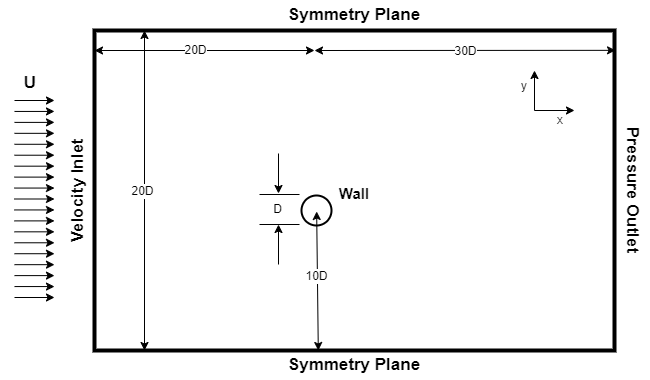}
     \caption{CFD Computational Domain }
     \label{fig: computational domain}
 \end{figure*}

\begin{figure*}[h!]
     \centering
     \includegraphics[width=105mm]{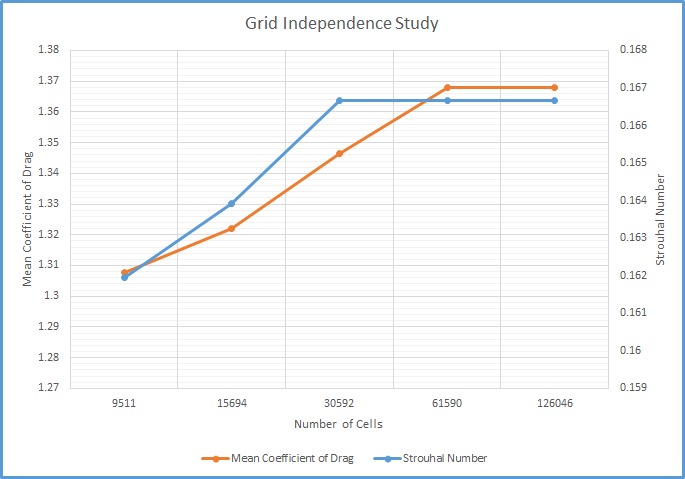}
     \caption{Grid Independence Study }
     \label{fig: grid}
 \end{figure*}

\begin{figure*}[h!]
     \centering
     \includegraphics[width=105mm]{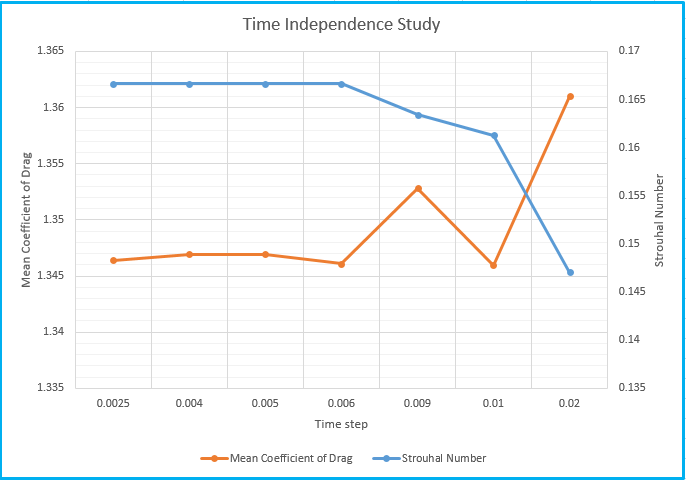}
     \caption{Time Independence Study }
     \label{fig: time}
 \end{figure*}

\addcontentsline{toc}{subsubsection}{1. CFD MODEL SETUP}
The training and test datasets for the case of 2D unsteady laminar flow past a cylinder are generated using STAR CCM+, a commercial CFD solver. The selected Reynolds number is 100. Fig. \ref{fig: computational domain} depicts the computational domain used for simulating the 2D unsteady laminar flow past a circular cylinder. Time and grid independence studies are performed to validate the simulation against the existing literature[\cite{silva2003numerical}]. Fig.~\ref{fig: grid} \& Fig.~\ref{fig: time}, respectively, show the grid and time-independent study results for the Strouhal number and the drag coefficient. The obtained Strouhal number and Mean Drag coefficient from the CFD simulation align closely with findings in the existing literature[\cite{silva2003numerical}], affirming the computational study's accuracy. As shown in  Fig.~\ref{fig: model training domain}, a rectangular domain is chosen in the wake of the circular cylinder for training the PINN model.
\subsubsection*{2. DATASET GENERATION}
\begin{figure*}[h!]
     \centering
     \includegraphics[width=\linewidth]{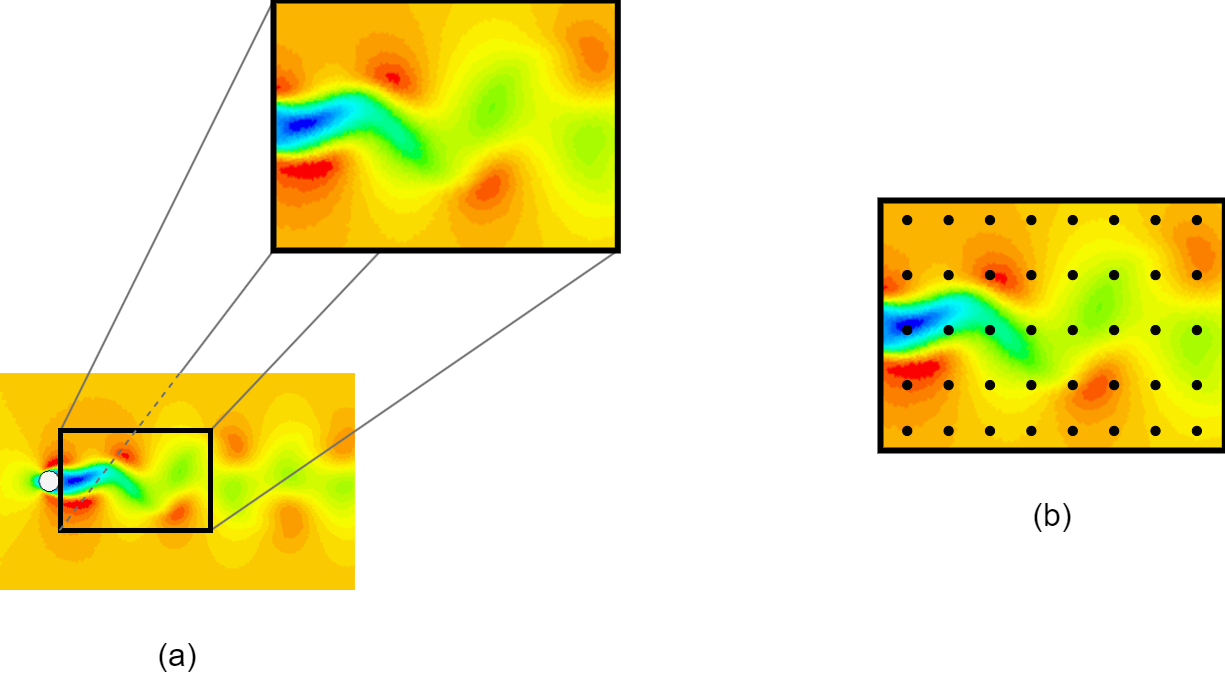}
     \caption{Illustration of (a) Surrogate Model Training domain (b) grid of spatial points for extracting velocity data}
     \label{fig: model training domain}
 \end{figure*}

\begin{figure*}[h!]
     \centering
     \includegraphics[width=\linewidth]{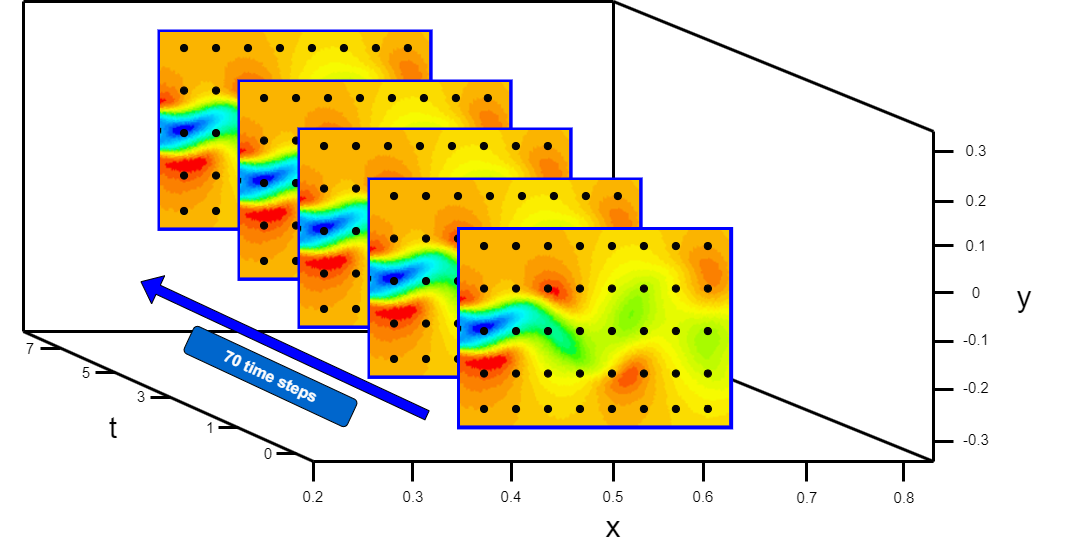}
     \caption{The space-time distribution of training dataset}
     \label{fig: train-space-time distribution}
 \end{figure*}

\begin{figure*}[h!]
     \centering
     \includegraphics[width=\linewidth]{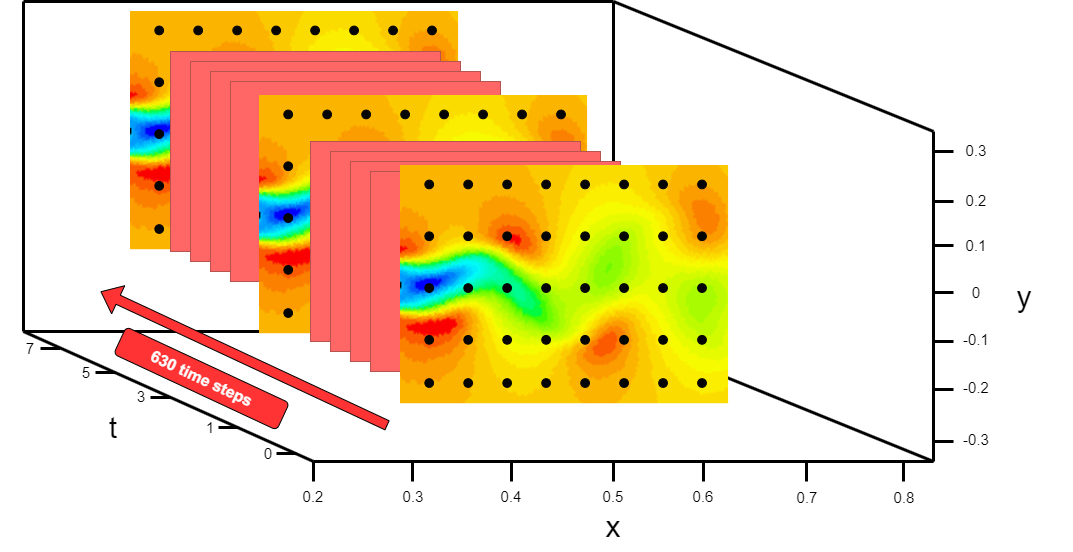}
     \caption{The space-time distribution of test dataset}
     \label{fig: test-space-time distribution}
 \end{figure*}

\addcontentsline{toc}{subsubsection}{2. DATASET GENERATION}
A rectangular domain is chosen in the wake of the cylinder as depicted in fig.~\ref{fig: model training domain}(a). Fig.~\ref{fig: model training domain}(b) illustrates the spatial distribution of flow field sample points. A grid of 100 x 100 points is chosen to extract velocity and pressure data. This data acts as ground truth data for training and testing purposes. It should be noted that the spatial domain for training all surrogate models, i.e., both data-driven and PINN models, is sparsely sampled with only 96 data points, which is less than 1\% of the total CFD collocation points. The trained model is tested on the rest of the 99\% data points which serve as the ground truth.  The sampling frequency for training these surrogate models is 10Hz and the temporal domain for training the surrogate model is [0s,7s] with a \textit{dt}=0.1s. Fig.~\ref{fig: train-space-time distribution} illustrates the space-time distribution of the training dataset, which comprises 70 time steps. To quantify the performance of the surrogate model, the model is tested on the test dataset comprising the unseen intermediate time steps, which comprise 630 time steps (with $dt=0.01$). Fig. \ref{fig: test-space-time distribution} illustrates the space-time distribution of the test dataset. The model is then tested on the unseen intermediate time steps marked in pink in fig.~\ref{fig: test-space-time distribution}. The relative $L_2$ error metric evaluates the performance of the model. The metric is defined at each time step in Eqn.~(\ref{relative l2})
\begin{equation}
    \epsilon_{u}=\frac {||{\hat{U}-U}||_2 }{||{U}||_2}
    \label{relative l2}
\end{equation}
where,
$\epsilon_v$ is the relative $L_2$ error,$\hat{U}$ denotes the predictions generated by the surrogate model, and $U$ refers to the ground truth obtained from the CFD simulation.

% It has to be noted that the surrogate model should not be confused with an autoregressive model for time series prediction. This surrogate model aims to reconstruct the complete spatial flow field from limited available data in the given temporal domain. 

\subsubsection*{3. COMPUTATIONAL RESOURCES}
\addcontentsline{toc}{subsubsection}{3. COMPUTATIONAL RESOURCES}
Unless otherwise specified, each training cycle is constrained to a computational budget of $2.5e04$ iterations with learning rates of $[1e-03,1e-04]$ and the size of the mini-batch used is $1e03$. The models are trained and evaluated on a single NVIDIA GeForce RTX 3060 GPU equipped with 12 GB of memory and 3584 CUDA cores.

\subsection*{B. DATA-DRIVEN MODELLING} \label{2.2}
\addcontentsline{toc}{subsection}{B. DATA-DRIVEN MODELLING}
For flow-related problems, a fully connected feed-forward neural network (FC-FFNN) is used as the main framework (ref.\cite{cai2021physics}), with the spatio-temporal coordinates as the inputs and the flow-field variables as the outputs of the network. In an FC-FFNN, the outputs are modeled as a nonlinear combination of the inputs and the parameters of the network, enabling the FC-FFNN to capture complex relationships and patterns within the flow-field variables.
This process is described as,
\begin{equation}
\textbf{h}0=(x,t)
\end{equation}
\begin{equation}
q_i=W_i\textbf{h}{i-1}+b_i
\end{equation}
\begin{equation}
h_i=\psi ( q_i), i=1,2,3,...,I, and
\end{equation}
\begin{equation}
\textbf{O}=W_{i+1}h_i+b_{i+1}
\label{nn_output}
\end{equation}

Here, $(x,t)$ $\epsilon$ $\mathbb{R}^{n_{x+1}}$ denotes the spatio-temporal input, while $h_0$ $\epsilon$ $\mathbb{R}^{n_{x+1}}$ represents the input layer, with $n_x$ as the spatial dimension. The preactivation unit of the network is denoted by $z_i$ $\epsilon$ $\mathbb{R}^{n_{i}}$, and the activation unit is denoted by $h_i$ $\epsilon$ $\mathbb{R}^{n_{i}}$. The neural network contains $I$ hidden layers, each with $n_i$ neurons, and $\psi (.)$ is the non-linear activation function. The weights and biases in the network are denoted as $W_i$ $\epsilon$ $\mathbb{R}^{n_{i-1}\times n_i}$ and $b_i$ $\epsilon$ $\mathbb{R}^{n_{i}}$ for $1 < i < L+1$. The network's weights and biases are collectively referred to as $\theta$. In eqn.~\ref{nn_output}, $\textbf{O}$ represents the output from the final layer of the network.

The network is trained on a dataset comprising inputs and their corresponding outputs, expressed as $[(x^j,t^j,\hat{o}^j]{j=0}^N{train}$, where $N_{train}$ is the total number of samples, and $\hat{o}$ denotes the ground truth. The training process involves minimizing a loss function $\textit{L}$ using a stochastic gradient descent (SGD)-based optimization algorithm:
\begin{equation}
\theta^{optimal}=\arg \min_{\theta} (\textit{L})
\end{equation}

The parameters $\theta$ are initialized either randomly or with the Xavier initialization scheme (ref.\cite{glorot2010understanding}) and updated iteratively during training. The SGD algorithm calculates the gradients $\nabla_\theta \textit{L}$ of the loss function $\textit{L}$ concerning the network parameters $\theta$ and updates them using the rule:
\begin{equation}
\theta_{n+1}=\theta_n - \eta \nabla_\theta \textit{L}(\theta_n)
\end{equation}

Here, $\eta$ represents the learning rate. Gradients are computed using backpropagation and automatic differentiation (AD) \cite{baydin2018automatic}. A mini-batch training strategy is employed, dividing the training data into smaller subsets to train the network iteratively. The parameter update rule for SGD, considering $\textit{P}$ mini-batches with $\textit{N}$ samples, is given by:
\begin{equation}
\theta_{n+1}= \theta_n - \eta_i * \frac{1}{N} \sum_{i^{\prime} =i_N+1}^{(i+1)N} \nabla \textit{L}_\theta(\theta,(x^{i^ {\prime} },t^{i^ {\prime} }), i=1,2,3,...P
\end{equation}

Here, $\eta_i$ is the learning rate for the $i^{th}$ epoch, with each epoch comprising $P=N_{train}/N$ SGD iterations. The ADAM optimizer \cite{kingma2014adam}, a variant of SGD, has been employed for optimization.

\subsubsection*{1. LOSS FUNCTION FORMULATION}
\addcontentsline{toc}{subsubsection}{1. LOSS FUNCTION FORMULATION}
To accurately replicate real-world scenarios, we randomly sample 96 data points within the spatial domain to emulate the sparsity often observed in real-world datasets. A fully connected feed-forward neural network is then trained on these datasets with the aim of minimizing the difference between the model predictions and ground truth. The loss function formulation for the data-driven models is as follows:
\begin{equation}
    Loss= MSE_u + MSE_v + MSE_p 
\end{equation}
where
\begin{equation}
    MSE_u=\frac{1}{N_q} \sum_{e=1}^{N_q} (\hat{u}(x_e^q,t_e^q)-u(x_e^q,t_e^q))^2
\end{equation}
\begin{equation}
    MSE_v=\frac{1}{N_q} \sum_{e=1}^{N_q} (\hat{v}(x_e^q,t_e^q)-v(x_e^q,t_e^q))^2-
\end{equation}
\begin{equation}
    MSE_p=\frac{1}{N_q} \sum_{e=1}^{N_q} (\hat{p}(x_e^q,t_e^q)-p(x_e^q,t_e^q))^2
\end{equation}

$\hat{u},\hat{v},\hat{p}$ are the model's predictions, while \(u\), \(v\), and \(p\) represent the ground truth data obtained from the CFD simulations. Each component of the loss function indicates the misfit between the model predictions and the ground truth. 

\subsubsection*{2. HYPER-PARAMETER TUNING}
\addcontentsline{toc}{subsubsection}{2. HYPER-PARAMETER TUNING}
Hyper-parameter tuning is performed to obtain the appropriate number of hidden layers and neurons per hidden layer for the data-driven models. The optimal configuration for the number of hidden layers and neurons per layer is determined through a grid search across combinations of $[5, 7, 8, 10]$ layers and $[50, 75, 100]$ neurons per layer. The models are trained on the training dataset comprising 96 data points per timestep, effectively utilizing only 1$\%$ of the available data. The resultant models are evaluated using the relative $L_2$ error metric defined in Eq.\ref{relative l2}, based on velocity reconstruction. Table \ref{hp data} shows the performance of models using different combinations of the hyperparameters.

Notably, the model configured with 5 hidden layers and 50 neurons per layer achieved the lowest average relative $L_2$ error(aRelative $L_2$ error), calculated as the mean of all relative $L_2$ errors across the test dataset, compared to other configurations.

\begin{table}[htbp]
\centering
\caption{Accuracy details of data-driven models trained on a dataset having 96 data points per time step for different model configurations.}
\begin{tabular}{p{3cm} p{3cm} p{3cm} p{3cm} }
\hline
Layers & Neurons & u-velocity aRelative $L_2$ error & pressure aRelative $L_2$ error \\
\hline
\multicolumn{1}{c}{\multirow{3}{*}{5}} & {50} & ${2.99\mathrm{e-}02}$& ${2.16\mathrm{e-}01}$ \\ 
\multicolumn{1}{c}{} & 75 & $5.43\mathrm{e-}02$& ${2.87\mathrm{e-}01}$ \\
\multicolumn{1}{c}{} & 100 & $5.52\mathrm{e-}02$& ${3.40\mathrm{e-}01}$ \\ \hline
\multicolumn{1}{c}{\multirow{3}{*}{7}} & 50 & ${3.56\mathrm{e-}02}$& ${2.12\mathrm{e-}01}$ \\
\multicolumn{1}{c}{} & 75 & $6.01\mathrm{e-}02$& ${3.42\mathrm{e-}01}$ \\ 
\multicolumn{1}{c}{} & 100 & $6.19\mathrm{e-}02$& ${3.73\mathrm{e-}01}$ \\ \hline
\multicolumn{1}{c}{\multirow{3}{*}{8}} & 50 & $3.62\mathrm{e-}02$& ${2.29\mathrm{e-}01}$ \\
\multicolumn{1}{c}{} & 75 & $5.96\mathrm{e-}02$& ${3.34\mathrm{e-}01}$ \\ 
\multicolumn{1}{c}{} & 100 & $7.71\mathrm{e-}02$& ${4.30\mathrm{e-}01}$ \\\hline
\multicolumn{1}{c}{\multirow{3}{*}{10}} & 50 & $5.59\mathrm{e-}02$& ${3.77\mathrm{e-}01}$ \\
\multicolumn{1}{c}{} & 75 & $5.52\mathrm{e-}02$& ${3.07\mathrm{e-}01}$ \\
\multicolumn{1}{c}{} & 100 & ${7.13\mathrm{e-}02}$& ${3.72\mathrm{e-}01}$ \\ \hline
\end{tabular}
\label{hp data}
\end{table}

\subsubsection*{3. PERFORMANCE ON TEST DATASET}
\begin{figure*}[h!]
     \centering
     \includegraphics[width=150 mm]{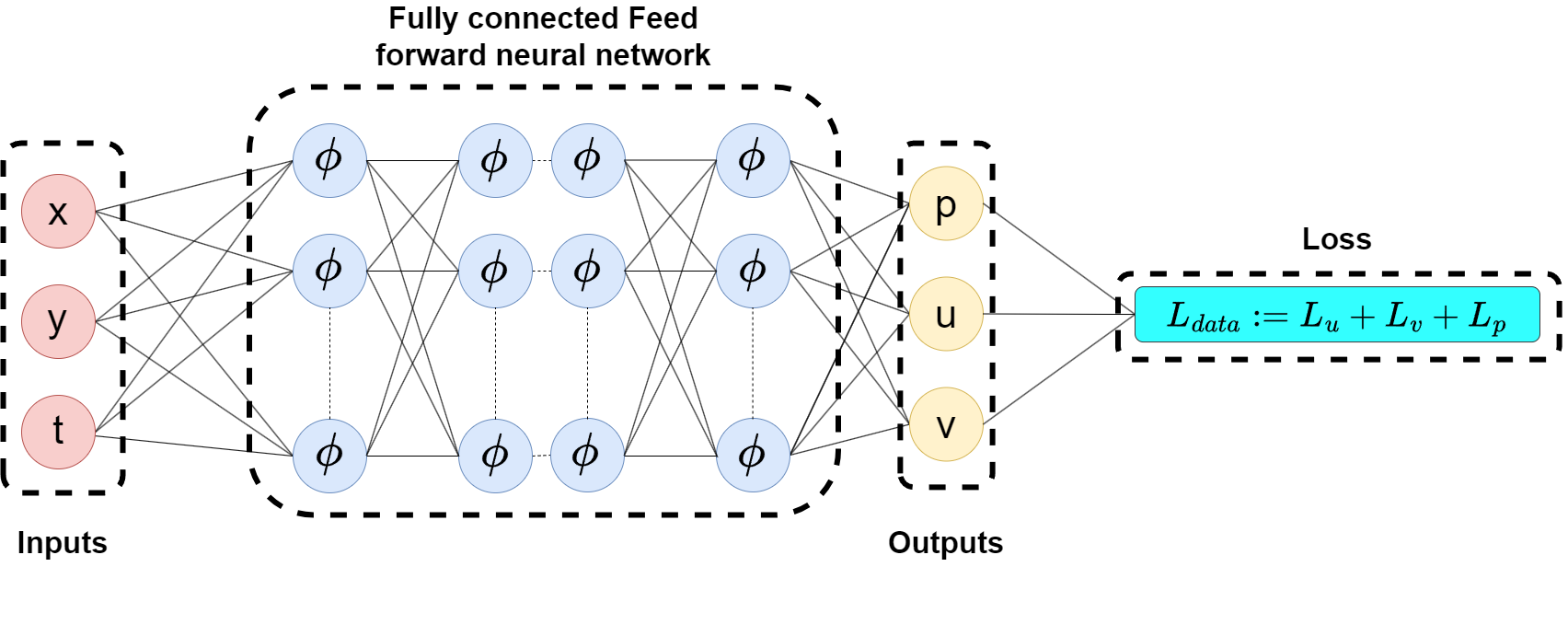}
     \caption{Schematic of Data-driven model}
     \label{fig: ddm-arch}
 \end{figure*}

\begin{figure*}[h!]
    \hspace{-3.5cm} \includegraphics[width=200mm]{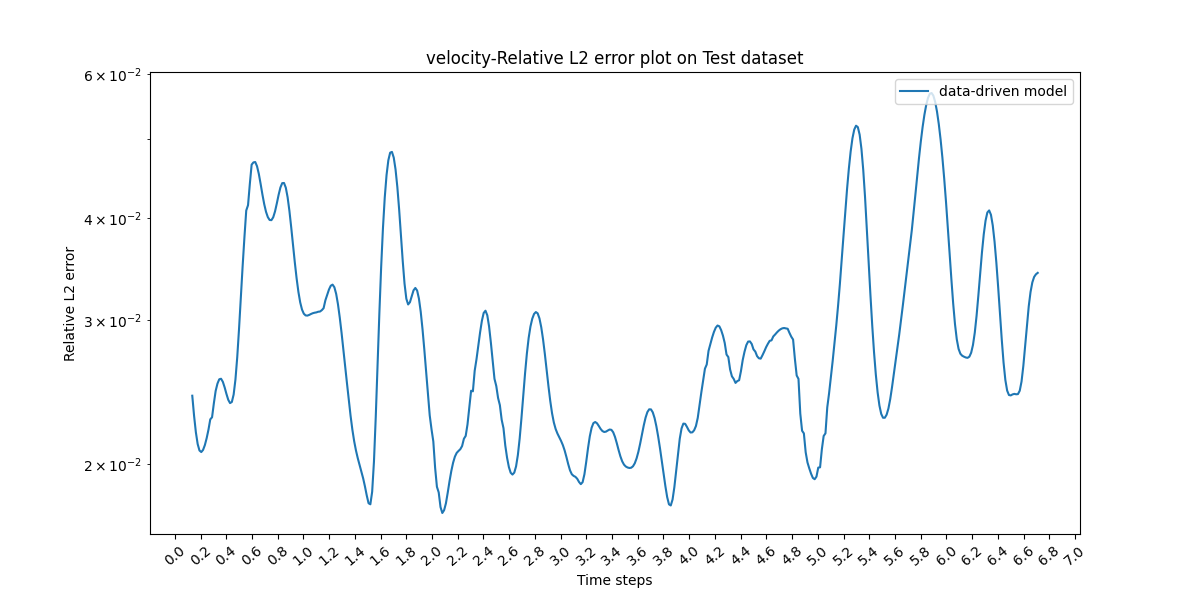}
     \caption{Performance of data-driven model on the u-velocity test dataset.}
     \label{fig: ddm-err}
\end{figure*}

\begin{figure*}[h!]
    \hspace{-3.5cm} \includegraphics[width=200mm]{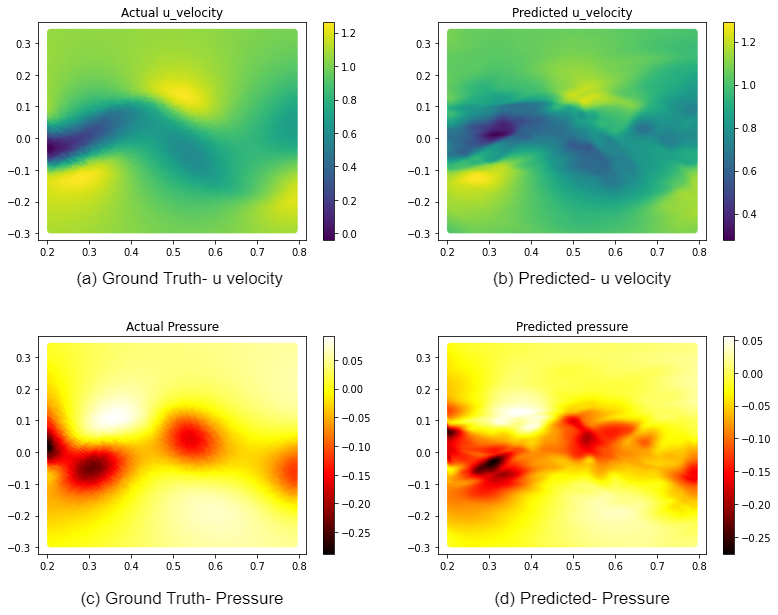}
     \caption{\textit{Top:}Comparison of true and predicted u-velocity (a and b)\\ \textit{Bottom:} Comparison of ground truth of pressure (c) and data-driven model's prediction of pressure (d) for a test time stamp.}
     \label{fig: ddm-preds}
\end{figure*}

\addcontentsline{toc}{subsubsection}{3. PERFORMANCE ON TEST DATASET}
The data-driven models utilize a fully connected feed-forward neural network architecture as depicted in fig.~\ref{fig: ddm-arch}. The inputs to the model are the spatio-temporal coordinates, and the outputs are u(x-component of velocity), v(y-component of velocity), and p(pressure). The model with the configuration obtained from hyper-parameter tuning is then evaluated on the test dataset. Fig.~\ref{fig: ddm-err} depicts the performance of the data-driven model on the test dataset, while Fig.~\ref{fig: ddm-preds} showcases the model's predictions at a specific test timestamp. Accurately reconstructing pressure is crucial for evaluating the model's understanding of underlying physics. The data-driven model's failure to do so, despite training with pressure values, highlights its inability to capture the underlying physics. This emphasizes the importance of physics-informed models to ensure predictions align with governing physical laws.
 .
 
\subsection*{C. PHYSICS INFORMED MODELLING} \label{2.3}
\addcontentsline{toc}{subsection}{C. PHYSICS INFORMED MODELLING}
The capacity of an FC-FFNN to learn the underlying patterns from data is very well documented. However, this ability largely depends on the volume of the training dataset, as previously discussed. In the limited availability of labeled datasets, data-driven models risk overfitting and predicting solutions inconsistent with the underlying physical laws. To this effect, the governing physical laws can be incorporated into the network (ref.\cite{karniadakis2021physics}). Their review paper presented three ways of integrating physical laws into neural networks.
\begin{enumerate}
    \item Observational bias, where the FC-FFNN is trained on large datasets with an aim to reduce the difference between the model predictions and ground truth. The FC-FFNN is expected to learn the underlying physics from the data during the training process.
    \item Inductive bias, where a specialized network architecture is constructed to incorporate the physical laws. This ensures that the model predictions inherently satisfy the governing physical laws.
    \item Learning bias, where the governing Partial Differential Equation (PDE) of the physical phenomenon is enforced as a penalty constraint in the loss function. Physics-informed neural networks (PINNs) fall under this bracket of incorporating physical laws into the neural networks.
\end{enumerate}

\subsubsection*{1. LOSS FUNCTION FORMULATION AND MODEL TRAINING}
\begin{figure*}[h!]
     
    \hspace{-2cm} \includegraphics[width=170mm]{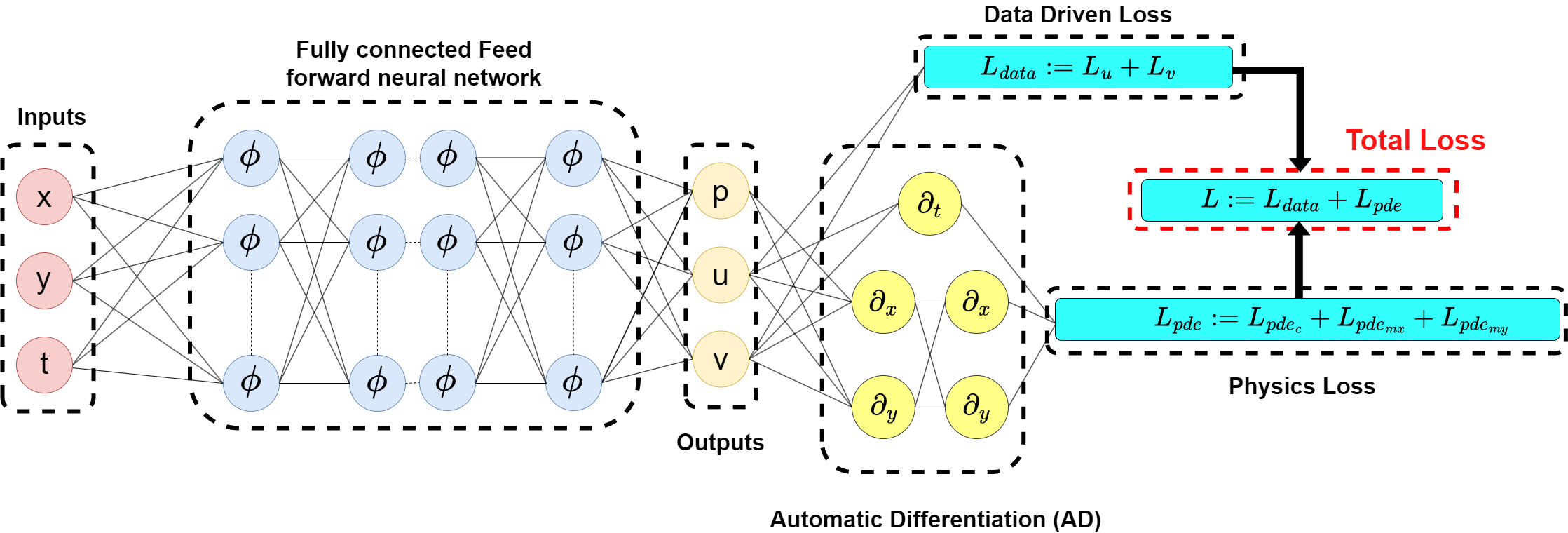}
     \caption{Schematic of Physics informed surrogate model}
     \label{fig: pinn-arch}
 \end{figure*}

\addcontentsline{toc}{subsubsection}{1. LOSS FUNCTION FORMULATION AND MODEL TRAINING}
PINNs, as formulated in the pioneering work of Ref.\cite{raissi2019physics}, aim to minimize a composite loss function given by Eqn.~(\ref{loss_eq}):
\begin{equation}
    L = L_{data} + L_{pde}
\label{loss_eq}
\end{equation}
 $L_{data}$ and $L_{pde}$ denote the loss contributions from the data component and the physics component, respectively. $L_{data}$ corresponds to the difference between the predictions of the model and the ground truth available at sparsely sampled wake field locations; $L_{pde}$ leverages the AD (Automatic Differentiation) of the TensorFlow framework \cite{abadi2016tensorflow} to calculate the gradients of the outputs $(u,v,p)$ with respect to the inputs $(x,y,t)$. These gradients are subsequently utilized to compute the residuals of the governing Partial Differential Equation (PDE) representing the flow, i.e., the Navier-Stokes Equations. The components of the composite loss function given by eqn.~(\ref{loss_eq}) are described as follows:
\begin{enumerate}
    \item $L_{pde}$: Residual loss component represents the mean squared error (MSE) loss arising from the residuals of the Navier-Stokes equations, which govern the system The following equations \ref{xcomp}-\ref{cont} correspond to the x momentum, y momentum, and continuity residuals, respectively.
    \begin{equation}
    L_{pde_{m_x}}=\frac{1}{N_r} \sum_{k=1}^{N_r} (e_{mx}(x_{res}^k,t^k))^2
    \label{xcomp}
    \end{equation}
    \begin{equation}
    L_{pde_{m_y}}=\frac{1}{N_r} \sum_{k=1}^{N_r} (e_{my}(x_{res}^k,t^k))^2
    \label{ycomp}
    \end{equation}
    \begin{equation}
    L_{pde_{c}}=\frac{1}{N_r} \sum_{k=1}^{N_r} (e_{c}(x_{res}^k,t^k))^2
    \label{cont}
    \end{equation}
    \begin{equation}
        L_{pde}=L_{pde_{m_x}}+L_{pde_{m_y}}+L_{pde_{c}}
    \label{phy_loss}
    \end{equation}
    
    \item $L_{data}$: Data loss represents the MSE loss estimated from the data constraints of the predictions and ground truth. 
    \begin{equation*}
     L_{u_d}=\frac{1}{N_d} \sum_{k=1}^{N_d} (\hat{u}(x_d^k,t_d^k)-u_d^k)^2
    \end{equation*}
    \begin{equation*}
     L_{v_d}=\frac{1}{N_d} \sum_{k=1}^{N_d} (\hat{v}(x_d^k,t_d^k)-v_d^k)^2
    \end{equation*}
    \begin{equation}
     L_{data}=L_{u_d}+L_{v_d}
    \label{data_loss}
    \end{equation}
    
\end{enumerate}
where $\hat u $ is the prediction PINN model, u is the ground truth, and $ N_r, N_d$ denotes the number of data points to calculate Residual loss and Data loss, respectively.
The residuals of the Navier-Stokes Equations are given as follows:
\begin{equation}
    e_{mc}(x_{res}^k,t^k)=\frac{\partial u}{\partial x}+\frac{\partial v}{\partial y}
\label{c_res}
\end{equation}
\begin{equation}
    e_{mx}(x_{res}^k,t^k)=\frac{\partial u}{\partial t}+\frac{\partial p}{\partial x} + u\frac{\partial u}{\partial x} + v\frac{\partial u}{\partial y}-\frac{1}{Re}(\frac{\partial^2 u}{\partial x^2}+\frac{\partial^2 u}{\partial y^2})
\label{x_res}
\end{equation}
\begin{equation}
    e_{my}(x_{res}^k,t^k)=\frac{\partial v}{\partial t}+\frac{\partial p}{\partial y} + u\frac{\partial v}{\partial x} + v\frac{\partial v}{\partial y}-\frac{1}{Re}(\frac{\partial^2 v}{\partial x^2}+\frac{\partial^2 v}{\partial y^2})
\label{y_res}
\end{equation}
where $(x_{res}^k,t^k)$ is the set of collocation points in the training domain, used to evaluate the residuals of the governing PDE for calculating the physics loss $L_{pde}$, and $(x_d^k,t_d^k)$ is the set of sparsely sampled data points in the domain, where the data loss $L_{data}$ component is evaluated. Fig.~\ref{fig: pinn-arch} illustrates the schematics of a physics-informed surrogate model.
 \\
In this work, we investigated two existing training methodologies of PINNs, Standard PINN formulation, as proposed in \cite{raissi2019physics} and Backward Compatible PINN (BC-PINN) formulation, as proposed in \cite{mattey2022novel}. 
\begin{enumerate}
    \item {Standard PINN formulation}: The standard PINN framework introduced in \cite{raissi2019physics} is designed to predict the complete spatiotemporal region simultaneously. This method involves selecting training points at random from the full spatiotemporal domain during each epoch.
    \item {Backward Compatible-PINN (BC-PINN) formulation}: In contrast to the standard PINN formulation, a novel approach is introduced in \cite{mattey2022novel}. This method divides the temporal domain into smaller time intervals, with a single neural network trained on each individual segment. The goal is to minimize the residuals of the PDE while ensuring the data constraints are met. During training, the model is penalized for any deviations from its predictions for the preceding time intervals. Fig.~\ref{fig: bcpinn_training} depicts the training process, where the red-highlighted section indicates the current time segment under training, and the yellow-highlighted sections represent the model's predictions for the preceding segments. To maintain continuity between consecutive time segments, the predictions of the model made for the final time step of the current time segment is chosen as the initial conditions for the next time segment (marked in blue). 
     \begin{figure*}[h!]
    \hspace{-2cm} \includegraphics[width=170mm]{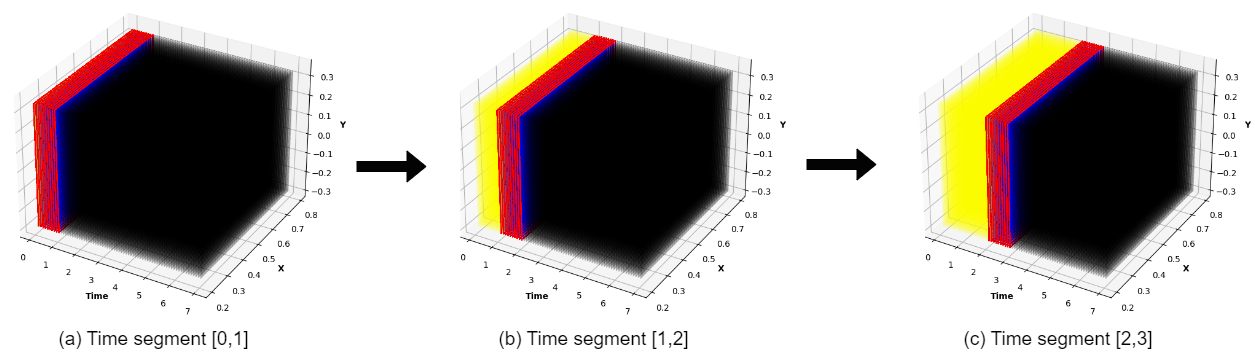}
     \caption{Schematic of BC-PINN training}
     \label{fig: bcpinn_training}
    \end{figure*}
\end{enumerate}
The loss function for BC-PINN formulation is as follows:
\begin{enumerate}
    \item First Time segment: 
        \begin{equation}
            L= L_{PDE}+L_{data}
        \end{equation}
    \item Consecutive Time segments:
    \begin{equation}
        L=L_{Prevdata}+L_{PDE}+L_{data}
    \end{equation}
\end{enumerate}
The loss term component $L_{Prevdata}$ ensures that the model is penalized for deviating from its own predictions for the previous time segments. 
\subsubsection*{2. HYPER-PARAMETER TUNING}
\addcontentsline{toc}{subsubsection}{2. HYPER-PARAMETER TUNING}
Hyper-parameter tuning is conducted by performing a grid search across the combinations of $[5,7,8,10]$ hidden layers and $[50,75,100]$ neurons per hidden layer. The physics-informed models are trained using Standard PINN formulation. Relative $L_2$ error is chosen as the performance metric, defined in Eq.~\ref{relative l2} based on the reconstruction of the stream-wise component of velocity. Table ~\ref{hp pinn} demonstrates the performance of models using different combinations of hyper-parameters. It can be seen from the table that the model with 10 hidden layers and 100 neurons per hidden layer was able to achieve the lowest relative $L_2$ error among all configurations. However, it is observed that the same configuration leads to overfitting in data-driven models, owing to higher error on the test dataset. For the network architecture comprising 3 inputs, 10 hidden layers with 100 neurons each, and 3 outputs, the total number of trainable parameters during the training process amounts to $8.1203 \times 10^4$. It can be seen that whilst utilizing the same data points, the physics-informed surrogate model is able to reduce the average relative u-velocity $L_2$ error by 6 times. Moreover, the accuracy in pressure prediction of the BC-PINN model is one order higher than the data-driven model. It should be noted that, unlike data-driven model, the PINN model does not use any direct information on the pressure data. The computational budget to perform hyper-parameter tuning is expanded to $1e05$ iterations with the learning rates of $[1e-03,1e-04]$ and a mini-batch size of $1e03$.

\begin{table}[htbp]
\centering
\caption{Accuracy details of physics-informed models trained on a dataset having 96 data points per time step for different model configurations.}
\begin{tabular}{p{3cm} p{3cm} p{3cm} p{3cm} }
\hline
Layers & Neurons & u-velocity  aRelative $L_2$ error & pressure aRelative $L_2$ error \\
\hline
\multicolumn{1}{c}{\multirow{3}{*}{5}} & 50 & $1.32\mathrm{e-}01$& $1.16\mathrm{e+}00$ \\ 
\multicolumn{1}{c}{} & 75 & $8.30\mathrm{e-}02$& $6.84\mathrm{e-}01$ \\
\multicolumn{1}{c}{} & 100 & $3.54\mathrm{e-}02$& $2.75\mathrm{e-}01$ \\ \hline
\multicolumn{1}{c}{\multirow{3}{*}{7}} & 50 & ${4.13\mathrm{e-}02}$& $3.08\mathrm{e-}01$ \\
\multicolumn{1}{c}{} & 75 & $2.59\mathrm{e-}02$& $1.91\mathrm{e-}01$ \\ 
\multicolumn{1}{c}{} & 100 & $2.01\mathrm{e-}02$& $1.35\mathrm{e-}01$ \\ \hline
\multicolumn{1}{c}{\multirow{3}{*}{8}} & 50 & $4.09\mathrm{e-}02$& $2.98\mathrm{e-}01$ \\
\multicolumn{1}{c}{} & 75 & $1.97\mathrm{e-}02$& $1.32\mathrm{e-}01$ \\ 
\multicolumn{1}{c}{} & 100 & $1.42\mathrm{e-}02$& $9.05\mathrm{e-}02$ \\\hline
\multicolumn{1}{c}{\multirow{3}{*}{{10}}} & 50 & $2.24\mathrm{e-}02$& $1.46\mathrm{e-}01$ \\
\multicolumn{1}{c}{} & 75 & $1.33\mathrm{e-}02$& $8.54\mathrm{e-}02$ \\
\multicolumn{1}{c}{} & 100 & $1.20\mathrm{e-}02$& $7.99\mathrm{e-}02$ \\ \hline
\end{tabular}
\label{hp pinn}
\end{table}

\subsubsection*{3. PERFORMANCE ON TEST DATASET}
\begin{figure*}[h!]
     \centering
     \includegraphics[width=\linewidth]{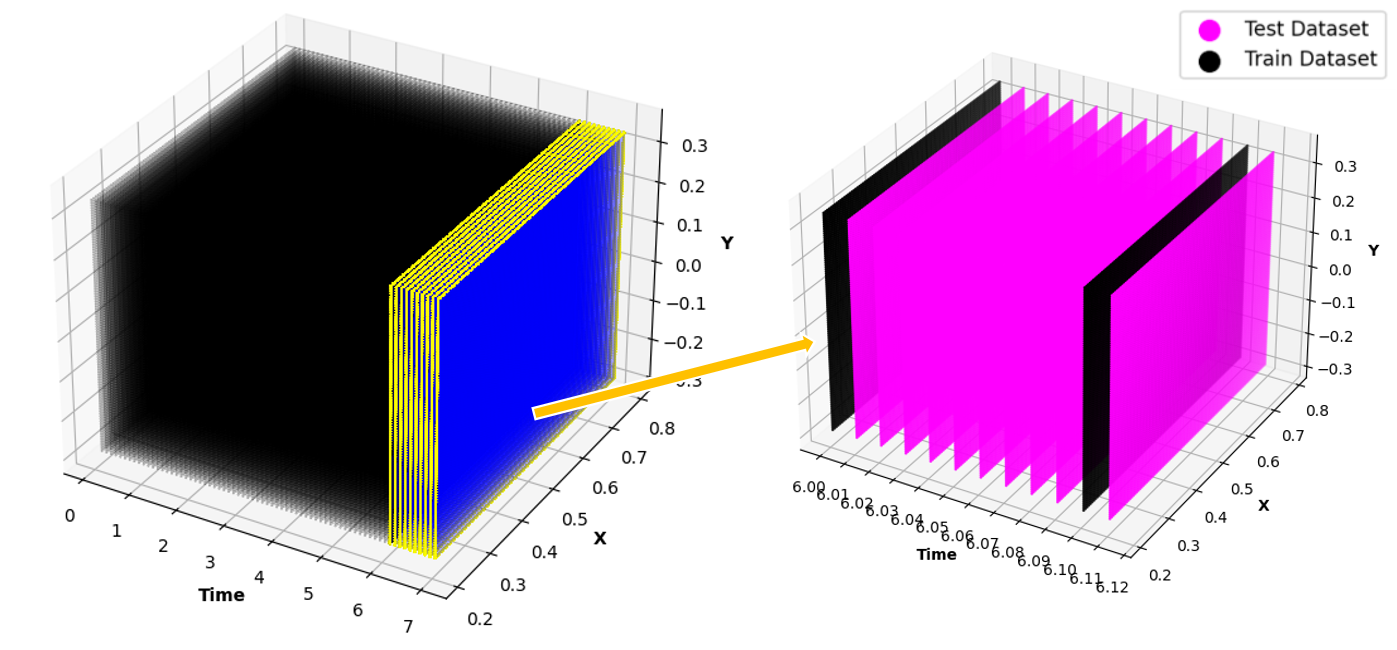}
     \caption{Training and testing dataset}
     \label{fig: Dataset}
 \end{figure*}

\begin{figure*}[h!]
    \hspace{-4cm} \includegraphics[width=200mm]{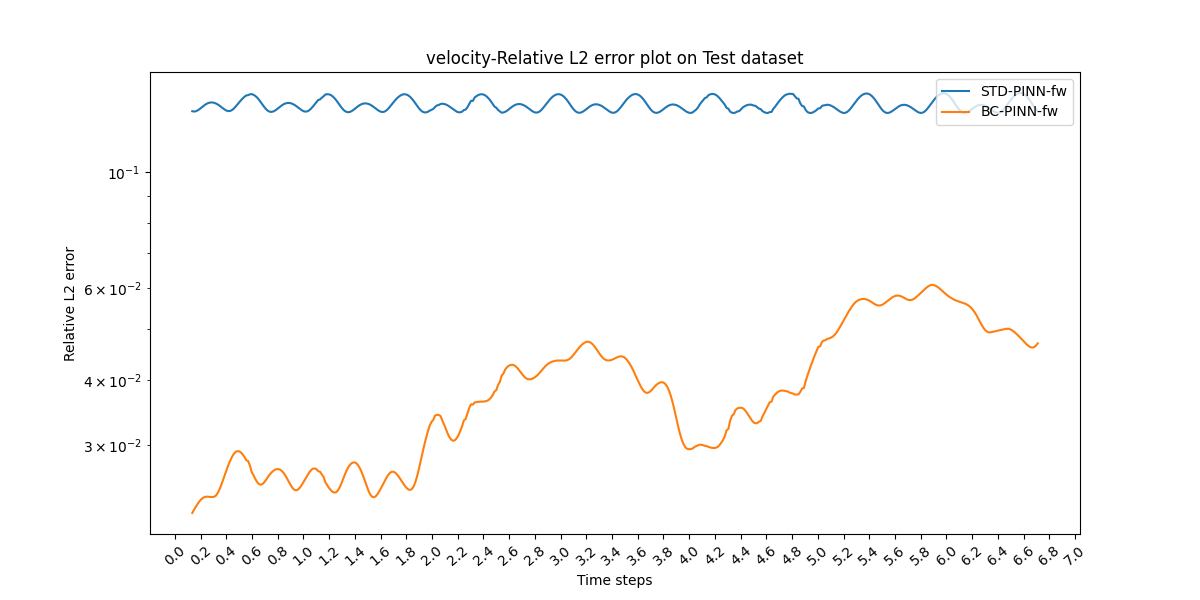}
     \caption{Performance of physics informed surrogate model using fixed weight strategy.}
     \label{fig: pinns-fw}
    \end{figure*}

\addcontentsline{toc}{subsubsection}{3. PERFORMANCE ON TEST DATASET}
The model with the optimal configuration obtained from hyper-parameter tuning is then evaluated on the test dataset, which comprises unseen intermediate time steps (marked in magenta in fig.~\ref{fig: Dataset}). Two surrogate models are trained using both the Standard PINN formulation and BC-PINN formulation, and their test performances are compared. Fig.~\ref{fig: pinns-fw} illustrates the performance of the models. The test performance of the models clearly demonstrates that the surrogate model trained using BC-PINN formulation for a comparable training duration outperforms the model trained using Standard PINN formulation. Further investigation is undertaken to gain insight into the underlying reasons for this phenomenon, and the findings are presented in the subsequent section.
 
\subsection*{D. UNDERSTANDING THE MULTI-OBJECTIVE OPTIMIZATION PROBLEM} \label{2.4}
\begin{figure*}[h!]
    \hspace{-4cm} \includegraphics[width=200mm]{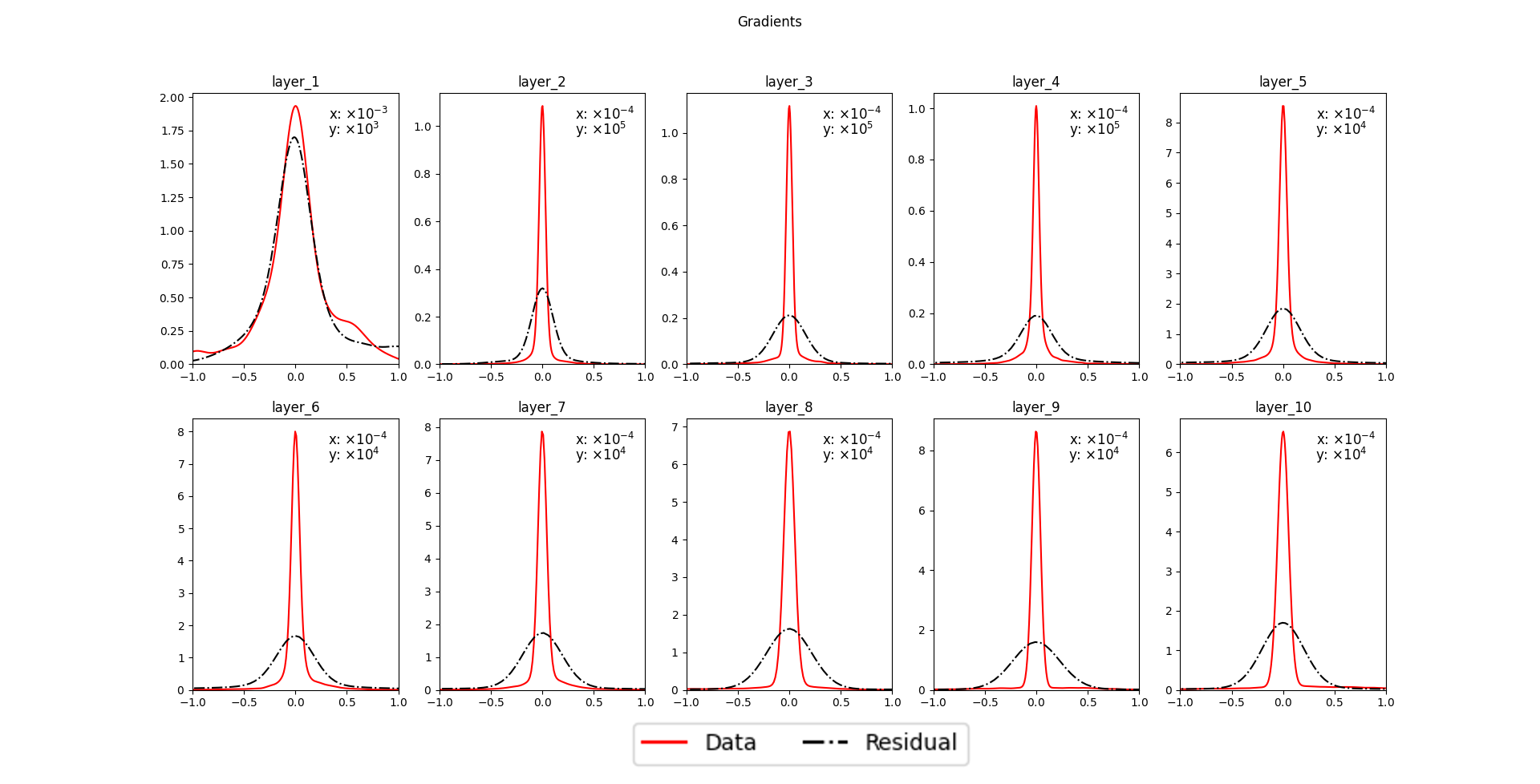}
     \caption{Histograms of back-propagated gradients of the surrogate model trained using Standard PINN formulation at the end of training}
     \label{fig: stdpinn-fw-grad}
    \end{figure*}

\begin{figure*}[h!]
    \hspace{-4cm} \includegraphics[width=200mm]{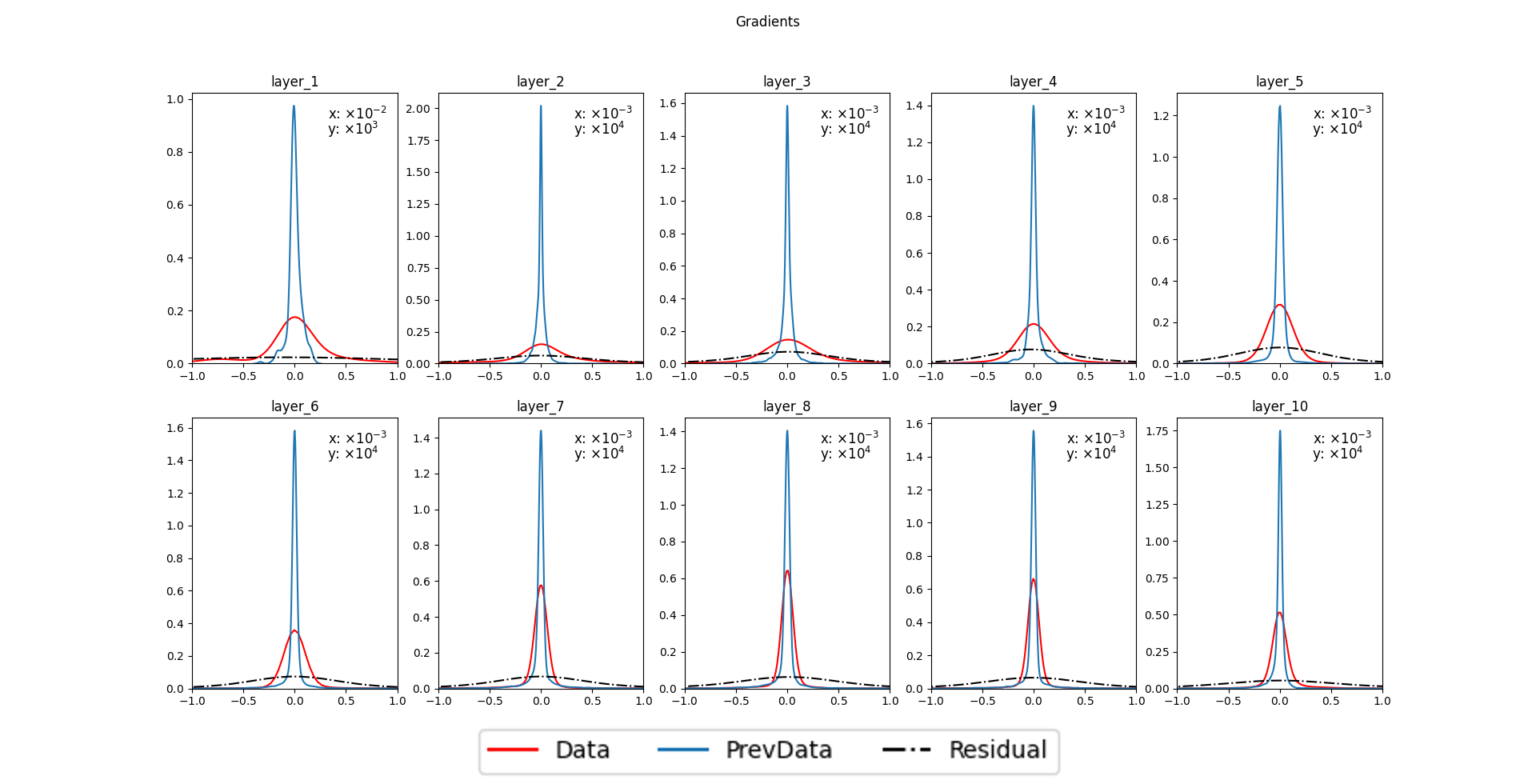}
     \caption{Histograms of back-propagated gradients of the surrogate model trained using BC-PINN formulation at the end of training}
     \label{fig: bcpinn-fw-grad}
    \end{figure*}

\addcontentsline{toc}{subsection}{D. UNDERSTANDING THE MULTI-OBJECTIVE OPTIMIZATION PROBLEM}
At its core, training a physics-informed surrogate model entails a multi-objective optimization problem. Here, the optimizer adjusts the model parameters to achieve a balance between minimizing the residual of the PDE and adhering to the data constraints. As a result, instead of observing the gradients of the entire loss function, the back-propagated gradients from each individual component of the loss function are monitored. Analyzing these gradients enables us to identify which component of the loss function has the most significant impact on the training process. Fig.~\ref{fig: stdpinn-fw-grad} and fig.~\ref{fig: bcpinn-fw-grad} display the histograms of the back-propagated gradients for the surrogate model trained using the Standard PINN and BC-PINN formulations, respectively, at the conclusion of the training, respectively. The x-axis denotes the gradient values, while the y-axis represents their corresponding frequency.

In the histograms depicted in fig.~\ref{fig: stdpinn-fw-grad} and fig.~\ref{fig: bcpinn-fw-grad}, the black dotted line represents the gradients flowing from the PDE residual term of the loss function, the red line corresponds to the gradients from the data loss component, and the blue line corresponds to the gradients from the previous time segment data loss component. It is evident from both figures that the back-propagated gradients flowing from the data loss term are sharply concentrated around zero, exhibiting significantly lower values compared to the gradients from the PDE residual term. 

The observed imbalance in gradients indicates that the model's training process is heavily skewed towards achieving a minimal residual of the PDE, possibly at the expense of effectively incorporating the available data constraints. Since a PDE has many solutions, it needs to be bounded by the data, which may not be happening when there is a gradient imbalance.  


To address this training imbalance, several strategies have been introduced in the existing literature. In \cite{sundar2024physics}, a systematic relaxation of physics loss is performed to understand the contribution of each loss function component to the training process. Recent studies (\cite{wang2021understanding},\cite{ jin2021nsfnets}) have introduced dynamic weighting strategies where the data-driven loss components are weighted. In this study, we initially examine the systematic relaxation of the PDE residual loss component and compare it with dynamic weighting strategies.

\subsubsection*{1. SYSTEMATIC RELAXATION OF PHYSICS LOSS COMPONENT}
\begin{figure*}[h!]
    \hspace{-4cm} \includegraphics[width=200mm]{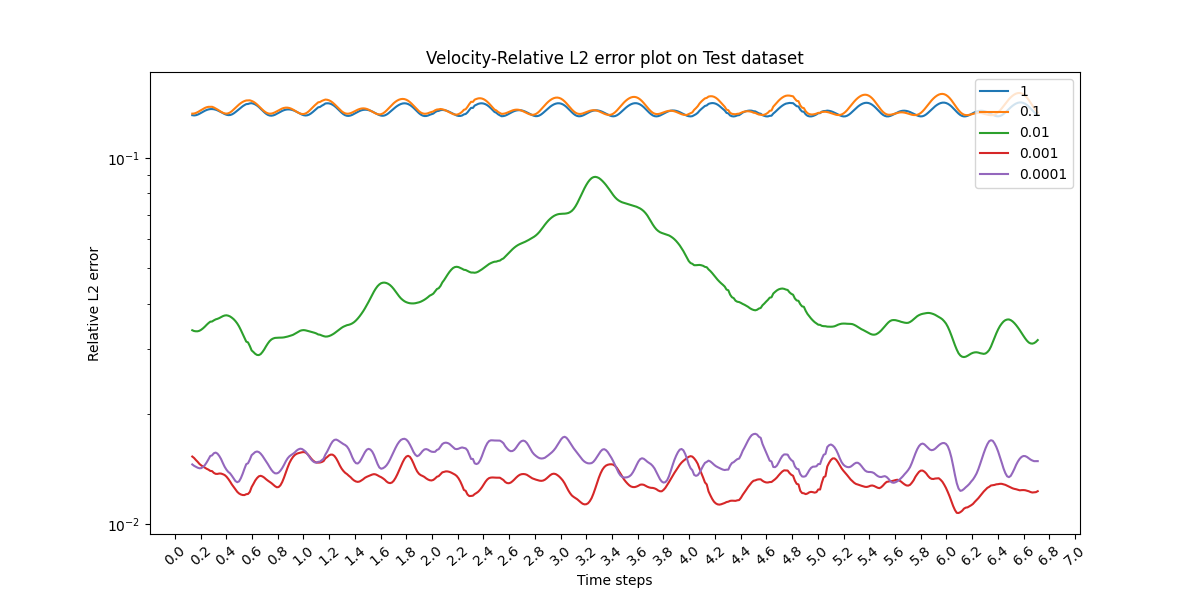}
     \caption{Performance of physics informed surrogate model using manually adjusted weights on test dataset for u-velocity }
     \label{fig: man_tune}
    \end{figure*}

\begin{figure*}[h!]
    \hspace{-4cm} \includegraphics[width=200mm]{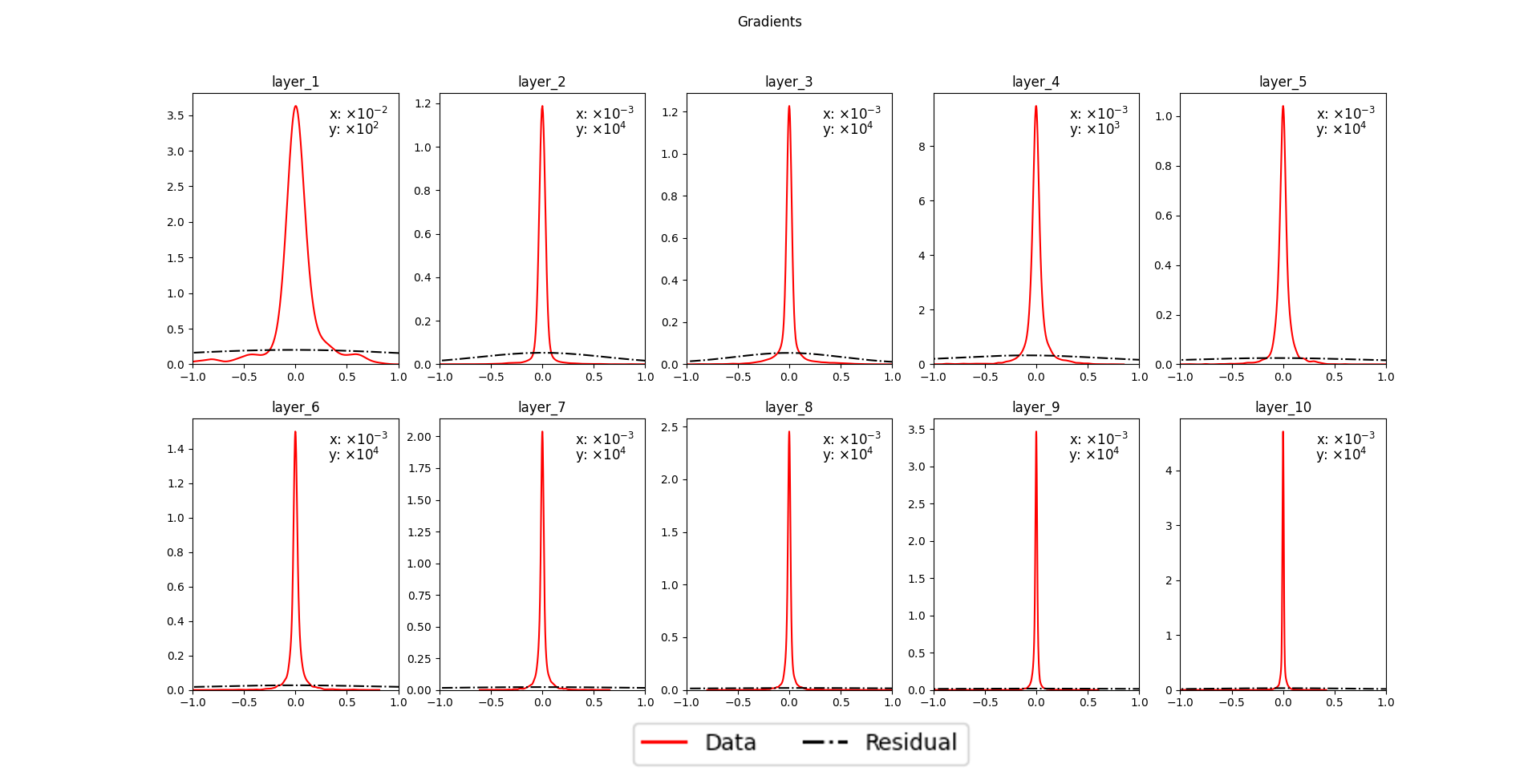}
     \caption{Histograms of back-propagated gradients of the surrogate model with 0.001 as the weighting constant}
     \label{fig: mantun-grad}
    \end{figure*}

\addcontentsline{toc}{subsubsection}{1. SYSTEMATIC RELAXATION OF PHYSICS LOSS COMPONENT}
In \cite{sundar2024physics}, the weightage given to the loss function component corresponding to the residual of the PDE is manually adjusted. The PDE residual loss component is relaxed by an order of 10, i.e. $[0.1,0.01,0.001,0.0001]$. A similar investigation is performed in this study, and histograms of the back-propagated gradients from each loss term component are examined after the training process. 
Physics-informed surrogate models are trained using Standard PINN formulation, which uses this systematic relaxation of the physics loss component, and the performance of these models on the test dataset is compared.
Table.~\ref{man_tab} presents the average Relative $L_2$ error obtained by different models trained using the given weighting constants for the PDE residual loss component. Fig.~\ref{fig: man_tune} illustrates the performance of different models on the test dataset.\\
It can observed with $0.001$ as the weighting constant for the PDE residual term, the least average relative $L_2$ error is achieved. Fig.~\ref{fig: mantun-grad} demonstrates the histograms of the back-propagated gradients at the end of the training. It is evident that the back-propagated gradients flowing from the data loss component (marked in red) remain sharply concentrated around zero, displaying significantly lower values compared to the gradients from the PDE residual term (marked in black).

\begin{table}[htbp]
\centering
\caption{Average Relative $L_2$ error for models with different weighting constants for PDE residual loss component}
\begin{tabular}{|c|c|}
\hline
\textbf{Weighting constant} & \textbf{aRelative $L_2$ error} \\
\hline
 1& $1.34\mathrm{e-}01$\\
 0.1& $1.36\mathrm{e-}01$\\
0.01& $4.43\mathrm{e-}02$\\
0.001&$1.32\mathrm{e-}02$\\
0.0001&$1.51\mathrm{e-}02$\\
\hline
\end{tabular}
\label{man_tab}
\end{table}

\subsubsection*{2. ADAPTIVE WEIGHTING}
\begin{figure*}[h!]
    \hspace{-4cm} \includegraphics[width=200mm]{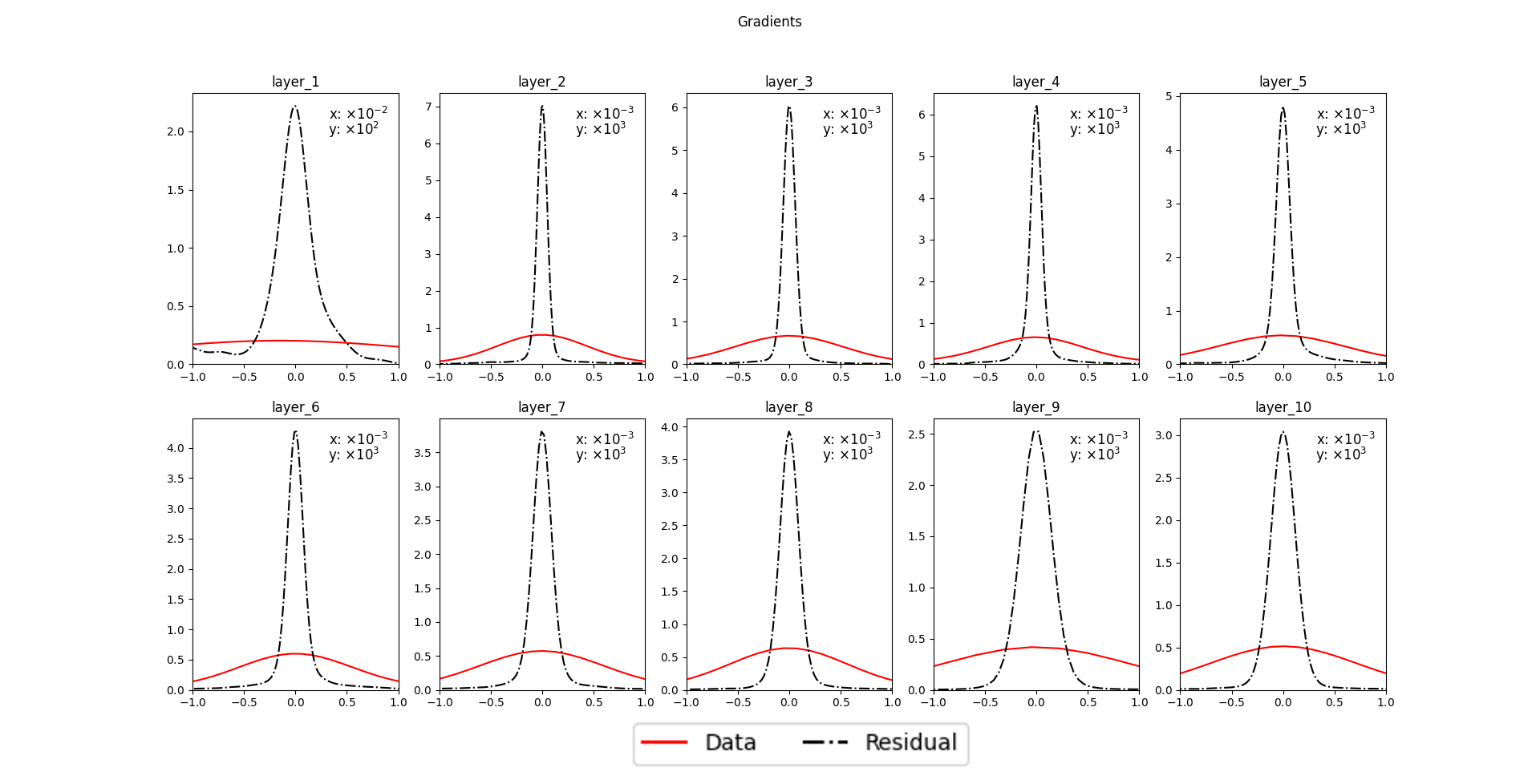}
     \caption{Histograms of back-propagated gradients of the surrogate model trained using Standard PINN formulation and dynamic weighting strategy}
     \label{fig: stdpinn-aw-grad}
    \end{figure*}

\begin{figure*}[h!]
    \hspace{-4cm} \includegraphics[width=200mm]{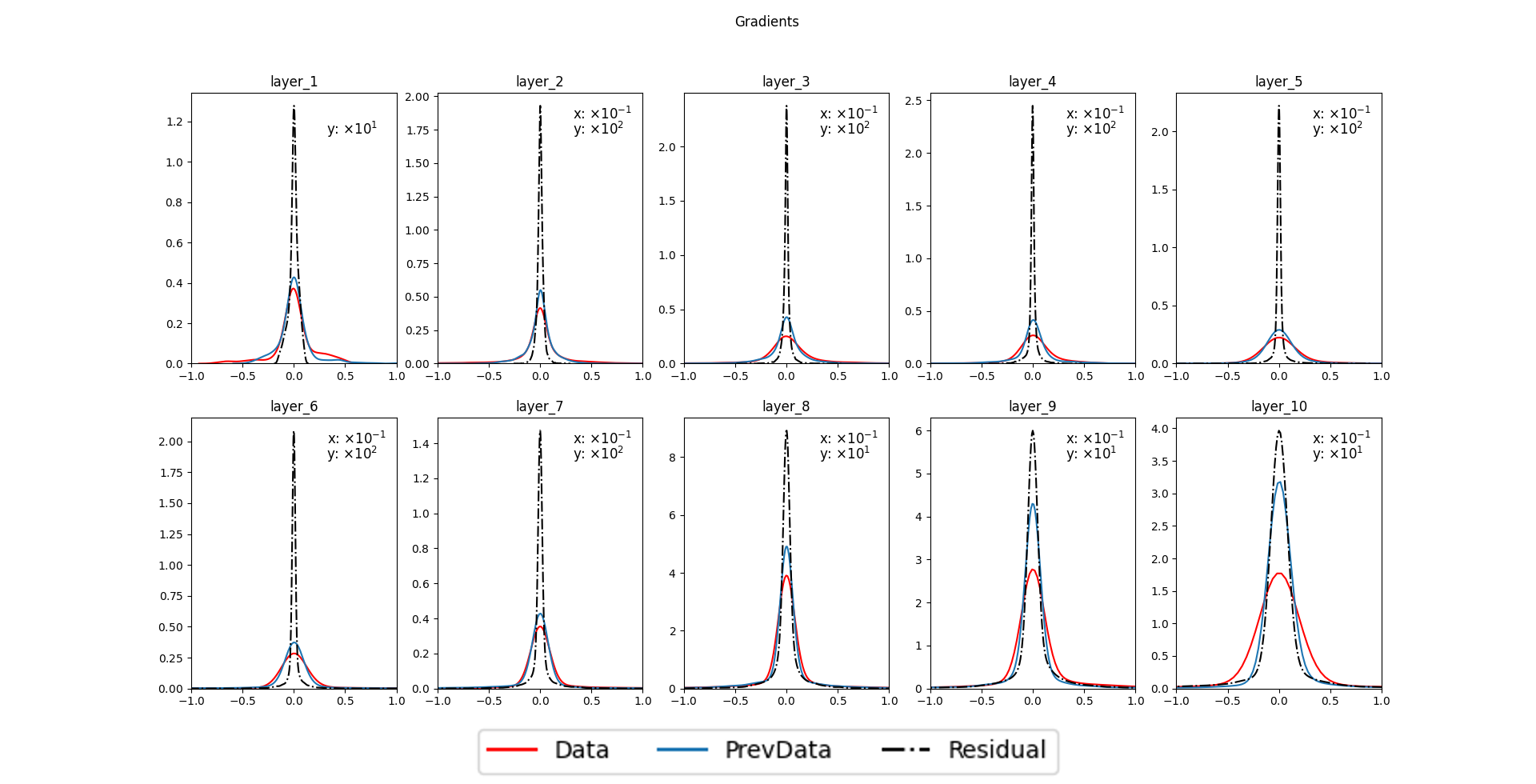}
     \caption{Histograms of back-propagated gradients of the surrogate model trained using BC-PINN formulation and dynamic weighting strategy}
     \label{fig: bcpinn-aw-grad}
    \end{figure*}

\addcontentsline{toc}{subsubsection}{2. ADAPTIVE WEIGHTING}
In ~\cite{wang2021understanding}, an adaptive weighting scheme was proposed where the weights of the data loss component of the loss function are calculated based on the statistics of the back-propagated gradients. This ensures that the interplay between the data loss term component and the PDE loss term component of the loss function is properly balanced. A similar adaptive weighting scheme was proposed in \cite{jin2021nsfnets}; however, in this study, we utilize the adaptive weighting scheme proposed in \cite{wang2021understanding}. In this method, first, the instantaneous values of the weighting constants are calculated using the ratio between the maximum gradient value obtained by $\nabla_\theta L_r(\theta)$, the gradient of the PDE residual loss, and the mean of the gradient magnitudes computed for the data-driven loss term, see eqn.\ref{eqn: inst-lamb}.
\begin{equation}
    \hat\lambda_d =\frac{\max_\theta (|\nabla_\theta L_r(\theta)|)}{\overline{\nabla_\theta L_d(\theta)}}
    \label{eqn: inst-lamb}
\end{equation}
\begin{equation}
    \lambda_d=(1-\alpha)\lambda_d+\alpha \hat \lambda_d
    \label{eqn: inst-avg}
\end{equation}
\begin{equation*}
    \alpha=0.9
\end{equation*}
where $\overline{\nabla_\theta L_d(\theta)}$ denotes the mean of $\nabla_\theta L_d(\theta)$ with respect to model parameters $\theta$. The actual dynamic weights are then calculated as a running average of the previous values, as shown in eqn.\ref{eqn: inst-avg}. Updates to these dynamic weights can occur either at every iteration of the training process or at any user-defined frequency.\\
Two surrogate models are trained using Standard PINN formulation and BC-PINN formulation using this adaptive weighting strategy; the histograms of the back-propagated gradients are then plotted after the training process. Fig.~\ref{fig: stdpinn-aw-grad} and fig.~\ref{fig: bcpinn-aw-grad} illustrate the histograms of the back-propagated gradients from the surrogate models trained using Standard PINN and BC-PINN formulation, respectively. The back-propagated gradients from the loss term component corresponding to the previous data loss are marked in blue.\\
It is evident from the histograms that the gradient balance between the two loss term components is better when the BC-PINN training methodology is used. In this study, we employ the BC-PINN formulation to train the surrogate models. 
    
\begin{figure*}[h!]
    \hspace{-4cm} \includegraphics[width=200mm]{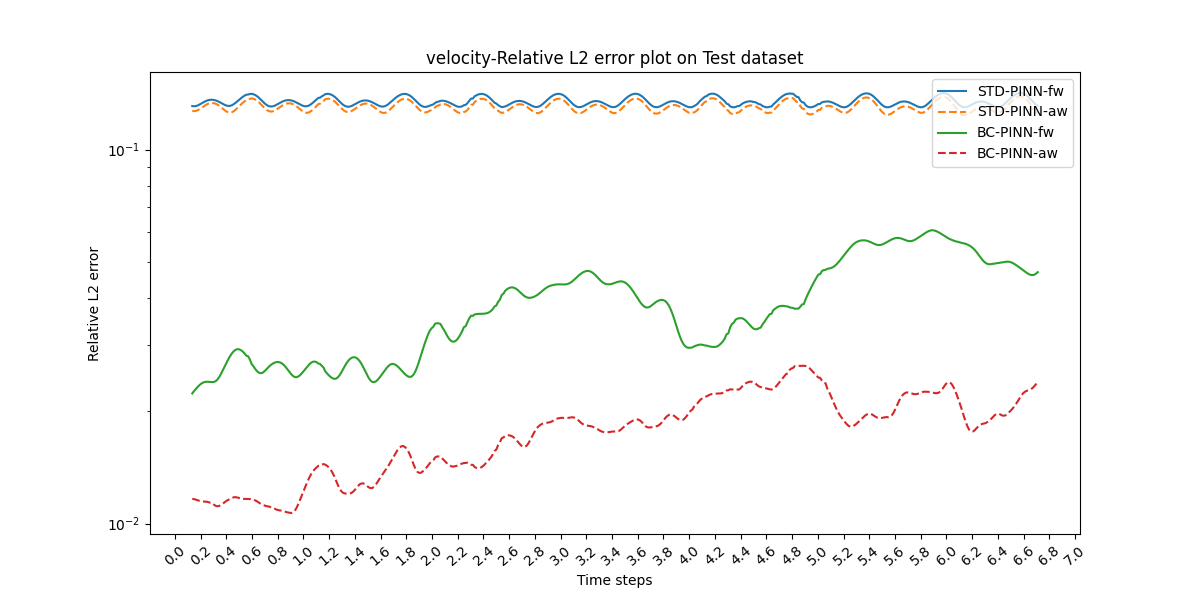}
     \caption{Performance of surrogate models using different weighting strategies on the test dataset for predicting u-velocity.}
     \label{fig: error_plots_std_bcp}
    \end{figure*}
    
\begin{figure*}[h!]
    \hspace{-3.5cm} \includegraphics[width=200mm]{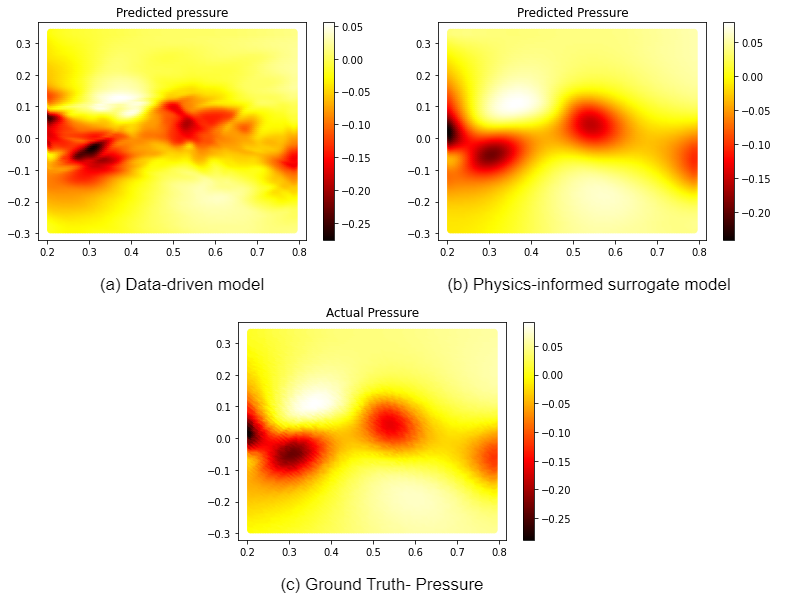}
     \caption{Illustrations of pressure predictions (a) Data-driven model (b) Physics-informed surrogate model and (c) Ground Truth for a test time stamp }
     \label{fig: pressure_preds}
    \end{figure*}

Fig.~\ref{fig: error_plots_std_bcp} illustrates the performance of the surrogate models using fixed weight strategy (marked in solid lines) and adaptive weighting strategy (marked in dotted lines). The error plot clearly demonstrates that employing an adaptive weighting strategy, as proposed in \cite{wang2021understanding}, results in the lowest error. \\

Fig.~\ref{fig: pressure_preds}(a) illustrates the pressure predictions made by the data-driven model fig.~\ref{fig: pressure_preds}(b) illustrates the pressure predictions made by the BC-PINN based physics-informed surrogate model. Notably, the physics-informed model is not trained using pressure data, yet it can recover pressure from sparse velocity datasets. This is in contrast to the data-driven model, which is trained on pressure data.

%%%%%%%%%%%%%%%%%%%%%%%%%%%%%%%%%%%%%%%%%%%%%%%%%%%%%%%%%%%%%%
%%%%%%%%%%%%%% 3D Turbulent flow past a ULCS %%%%%%%%%%%%%%%%%
%%%%%%%%%%%%%%%%%%%%%%%%%%%%%%%%%%%%%%%%%%%%%%%%%%%%%%%%%%%%%%

\section*{III. 3D TURBULENT FLOW IN THE WAKE OF AN ULTRA LARGE CONTAINER SHIP (ULCS)}
\label{ship}
\addcontentsline{toc}{section}{III. 3D TURBULENT FLOW IN THE WAKE OF AN ULTRA LARGE CONTAINER SHIP (ULCS)}

\subsection*{A. LABEL DATASET ACQUISITION AND DATASET GENERATION}
\addcontentsline{toc}{subsection}{A. LABEL DATASET ACQUISITION AND DATASET GENERATION}

\subsubsection*{1. CFD MODEL SETUP}
\begin{figure*}[h!]
\hspace{-4cm} \includegraphics[width=200mm]{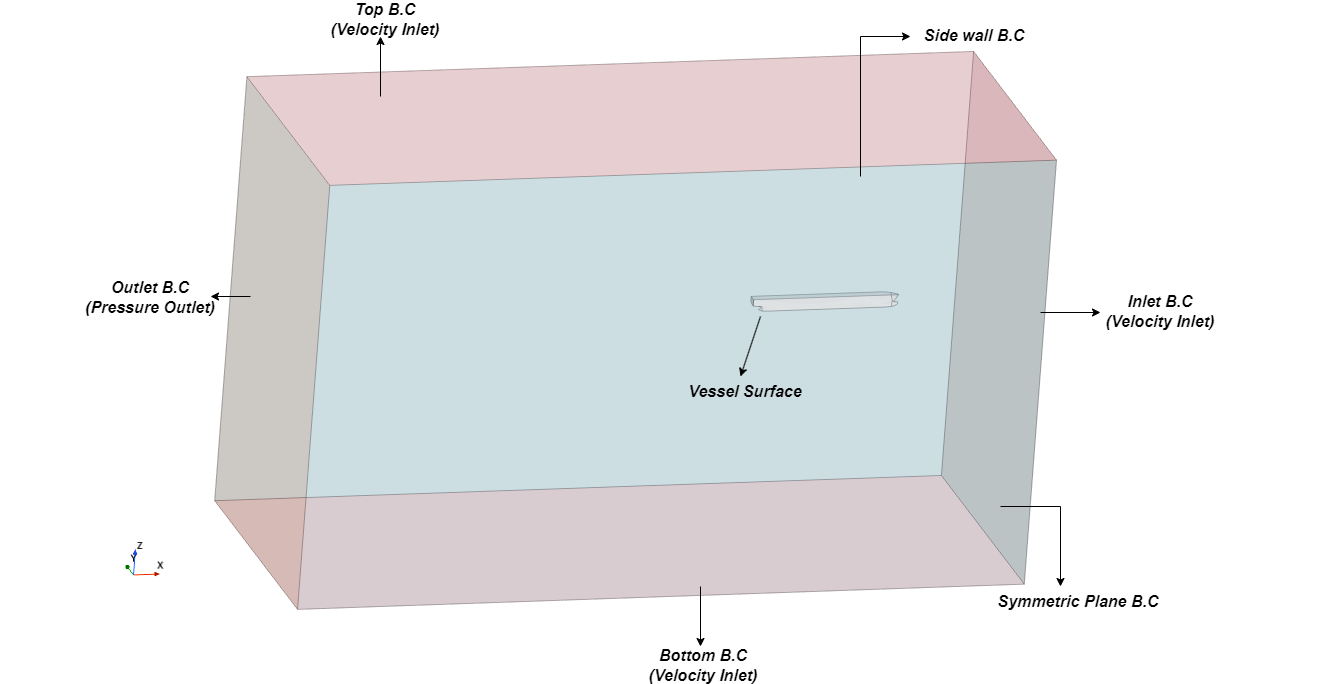}
 \caption{Illustration of Boundary conditions for the CFD simulation of the vessel}
 \label{fig: ship_boundary}
\end{figure*}

\begin{figure*}[h!]
\hspace{-2cm} \includegraphics[width=150mm]{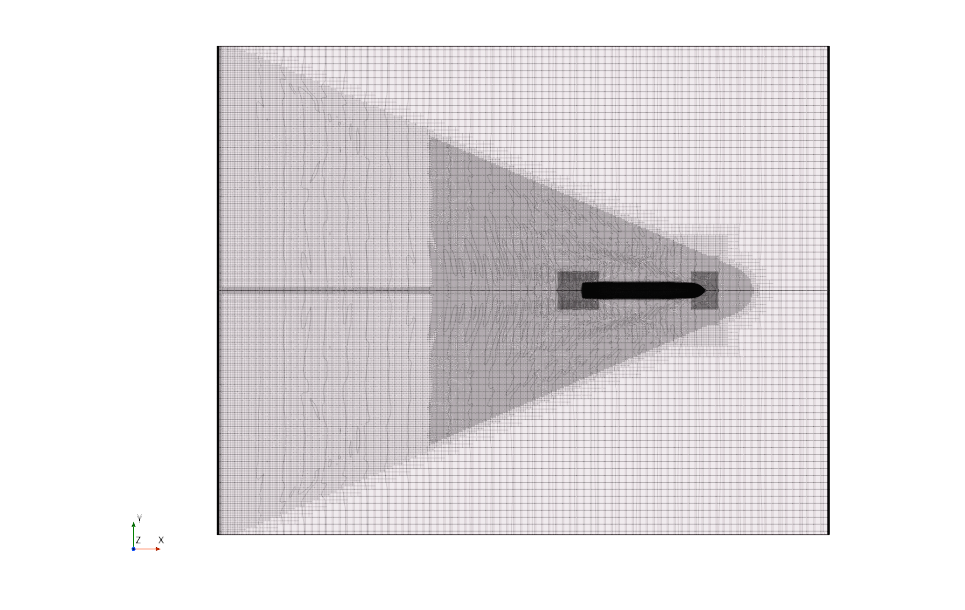}
 \caption{Illustration of 2D view of meshing around the vessel}
 \label{fig: ship_mesh}
\end{figure*}

\addcontentsline{toc}{subsubsection}{1. CFD MODEL SETUP}
The CFD simulation of a 3D unsteady turbulent flow past a ship is performed in Star CCM+, a commercial CFD solver. The parameters for the CFD solver are given in the table (Tab.~\ref{ship_cfd_details}). Fig.~\ref{fig: ship_boundary} illustrates the boundary conditions used for the CFD simulation of 3D unsteady turbulent flow past a ship. The boundary conditions for the simulation domain are detailed in tab.~\ref{ship_bound_details}. Fig.~\ref{fig: ship_mesh} provides an illustration of 2D view of meshing around the model of the vessel. The vessel's particulars used for the simulation are presented in tab.~\ref{ship_details}. In the computational domain, a trimmer cell technique is employed to create the volume mesh, while prism layers are incorporated to capture the boundary layer phenomena effectively. The volume mesh comprises a total of 4,592,220 cells, ensuring a finely discretized representation of the domain. Additionally, a y+ value of 75 is utilized in the simulation to resolve the near-wall flow characteristics appropriately.
\begin{table}[h]
\centering
\caption{CFD solver parameters for 3D unsteady turbulent flow past a ship}
\begin{tabular}{|c|c|}
\hline
\textbf{Parameter} & \textbf{Setting} \\
\hline
 Solver& 3D Stationary, Unsteady, Implicit \\
 Turbulence Model & Realisable $k-\epsilon$ \\ 
 Wall treatment & Two-layer all wall y+ treatment\\
 Multiphase flow model & Volume of Fluid (VOF), Gravity \\ 
 Pressure discretization & Standard \\ 
 Momentum discretization & Second order upwind \\
 Time discretization & First order upwind \\ 
 Turbulent kinetic energy discretization & Second order upwind \\
 Turbulence dissipation rate & Second order upwind \\ 
 Pressure-Velocity coupling & SIMPLE \\
\hline
\end{tabular}
\label{ship_cfd_details}
\end{table}

\begin{table}[h]
\centering
\caption{Boundary Conditions for the CFD simulation of 3d unsteady turbulent flow past a ship}
\begin{tabular}{|c|c|}
\hline
\textbf{Surface} & \textbf{Boundary condition} \\
\hline
 Vessel surface& Wall (no slip) \\
 Domain inlet, Top and Bottom & Velocity inlet\\
 Domain outlet & Pressure outlet\\
 Domain vertical surface at vessel symmetry & Symmetry plane\\ 
\hline
\end{tabular}
\label{ship_bound_details}
\end{table}

\begin{table}[h]
\centering
\caption{Principal dimensions of the vessel}
\begin{tabular}{|c|c|c|}
\hline
\textbf{Particulars} & \textbf{Prototype}& \textbf{Model} \\
\hline
 Scale & 1& 80 \\
 LOA & 122.25m& 5.25m \\
 LPP & 333.43m& 4.16m \\
 Beam & 42.8m& 0.525m \\
 Design Draft & 13.1m& 0.165m \\
 Displacement& 125000 t& 240.17 \\
 Speed & 15 knots& 0.862 m/s \\
 $C_B$& 0.66& 0.66 \\
 $C_P$ & 0.68& 0.68 \\
 
\hline
\end{tabular}
\label{ship_details}
\end{table}

\subsubsection*{2. TRAINING AND TESTING DATASETS}
\begin{figure*}[h!]
\hspace{-1cm} \includegraphics[width=150mm]{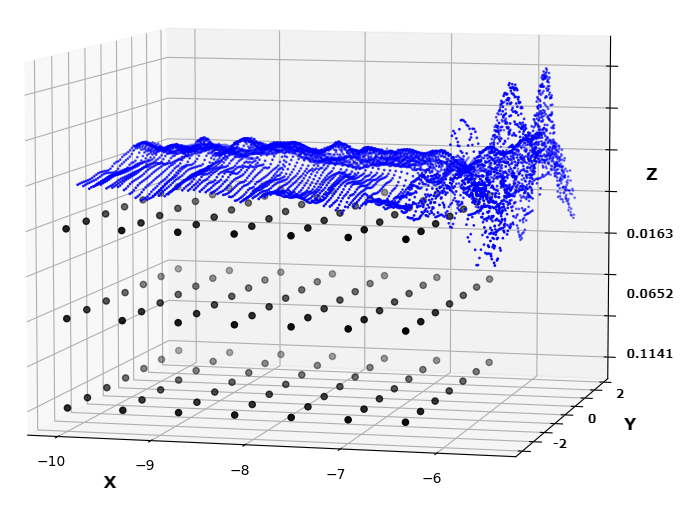}
 \caption{Illustration of Training domain in 3D}
 \label{fig: ship_grid}
\end{figure*}

\begin{figure*}[h!]
\hspace{1cm} \includegraphics[width=100mm]{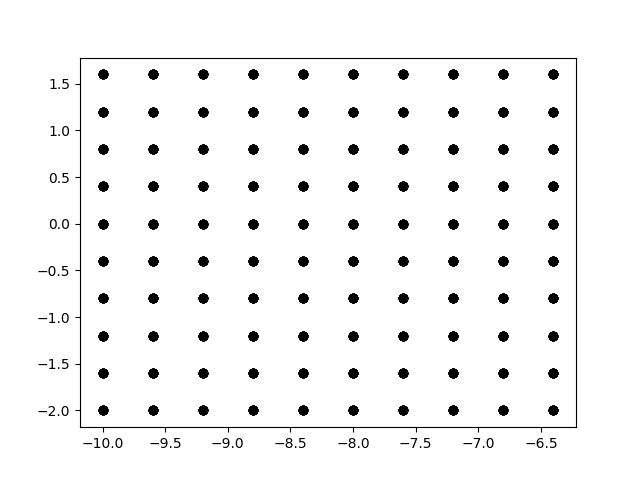}
 \caption{Grid of data points in 2D}
 \label{fig: ship_data}
\end{figure*}

\addcontentsline{toc}{subsubsection}{2. TRAINING AND TESTING DATASETS}
A 3D unsteady turbulent flow past a ship hull is simulated using Star CCM+, a commercial  Reynolds-Averaged Navier-Stokes (RANS) based CFD solver. Three grids are positioned in the wake region below the free surface (marked in blue) at specific depths, as illustrated in Fig.~\ref{fig: ship_grid}. Training and testing datasets are generated by collecting the data from these sparsely sampled points in the wake region. The temporal domain chosen for training the model is $[0s,7s]$ with a $dt=0.1s$ after the convergence is achieved. The model is subsequently evaluated on the test dataset, which includes unseen intermediate time steps, mirroring the methodology outlined in Section 2. Relative $L_2$ error(eqn.\ref{relative l2}) is used as the performance metric.
Fig.~\ref{fig: ship_data} illustrates a 2D view of the grid used to extract velocity data for generating the training and the testing dataset. 

\subsection*{B. NETWORK ARCHITECTURE AND LOSS FUNCTION FORMULATION}
\begin{figure*}[h!]
\hspace{-3.5cm} \includegraphics[width=200mm]{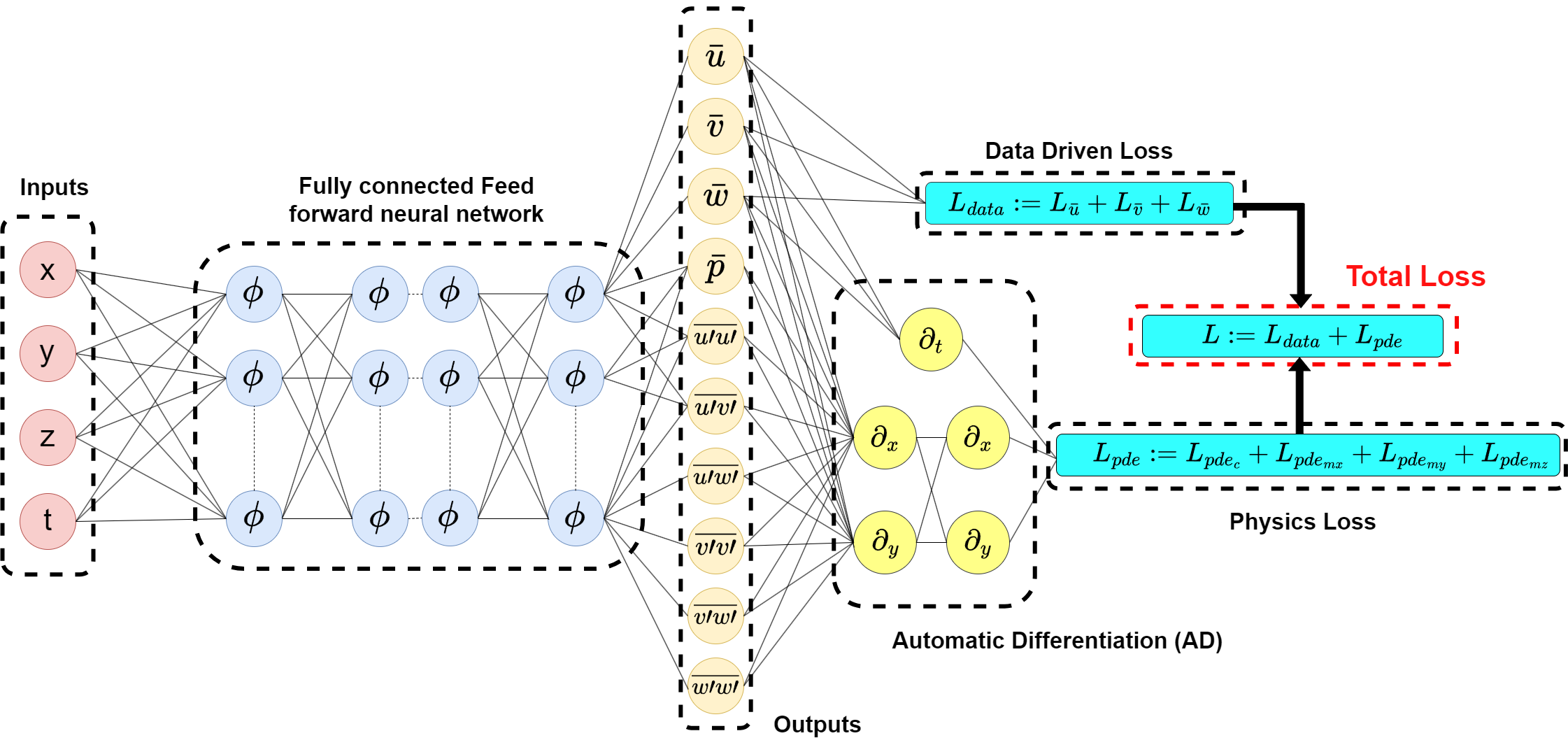}
 \caption{Schematic of Physics Informed Surrogate model for a 3D unsteady turbulent flow past a ship.}
 \label{fig: ship_pinn}
\end{figure*}

\addcontentsline{toc}{subsection}{B. NETWORK ARCHITECTURE AND LOSS FUNCTION FORMULATION}
As mentioned in the preceding section, adjustments to the network architecture are necessary since the governing Partial Differential Equation (PDE) for the data is the RANS equation. Fig. \ref{fig: ship_pinn} illustrates the schematic used for the physics-informed surrogate model. The inputs to the model are the spatiotemporal coordinates $(x,y,z,t)$, and the outputs are the  mean components of velocity $(\Bar{u},\Bar{v},\Bar{w})$, pressure $\bar{P}$, and Reynolds-stress components $(\overline{u\prime^2},\overline{v\prime^2},\overline{w\prime^2},\overline{u\prime v\prime},\overline{u\prime w\prime},\overline{v\prime w\prime},\overline{v\prime v\prime},\overline{v\prime w\prime},\overline{w\prime w\prime})$.
For a 3D incompressible unsteady turbulent fluid flow, the RANS equations are described as follows:
\begin{equation}
    \frac{\partial \bar u}{\partial x}+\frac{\partial \bar v}{\partial y}+\frac{\partial \bar w}{\partial z}=0
\label{conti_r}
\end{equation}
\begin{equation}
    \frac{\partial \bar u}{\partial t}+\frac{\partial \bar p}{\partial x}+\bar u\frac{\partial \bar u}{\partial x}+\bar v\frac{\partial \bar u}{\partial y}+\bar w\frac{\partial \bar u}{\partial z}-\frac{1}{Re}(\frac{\partial^2 \bar u}{\partial x^2}+\frac{\partial^2 \bar u}{\partial y^2}+\frac{\partial^2 \bar u}{\partial z^2})+\frac{\partial \overline{u\prime u\prime}}{\partial x}+\frac{\partial \overline{u\prime v\prime}}{\partial y}+\frac{\partial \overline{u\prime w\prime}}{\partial z}=0
\label{xmoment_r}
\end{equation}
\begin{equation}
    \frac{\partial \bar v}{\partial t}+\frac{\partial \bar p}{\partial y}+\bar u\frac{\partial \bar v}{\partial x}+\bar v\frac{\partial \bar v}{\partial y}+\bar w\frac{\partial \bar v}{\partial z}-\frac{1}{Re}(\frac{\partial^2 \bar v}{\partial x^2}+\frac{\partial^2 \bar v}{\partial y^2}+\frac{\partial^2 \bar v}{\partial z^2})+\frac{\partial \overline{v\prime u\prime }}{\partial x}+\frac{\partial \overline{v\prime v\prime }}{\partial y}+\frac{\partial \overline{v\prime w\prime }}{\partial z}=0
\label{ymoment_r}
\end{equation}
\begin{equation}
    \frac{\partial \bar w}{\partial t}+\frac{\partial \bar p}{\partial z}+\bar u\frac{\partial \bar w}{\partial x}+\bar v\frac{\partial \bar w}{\partial y}+\bar w\frac{\partial \bar w}{\partial z}-\frac{1}{Re}(\frac{\partial^2 \bar w}{\partial x^2}+\frac{\partial^2 \bar w}{\partial y^2}+\frac{\partial^2 \bar w}{\partial z^2})+\frac{\partial \overline{w\prime u\prime }}{\partial x}+\frac{\partial \overline{w\prime v\prime }}{\partial y}+\frac{\partial \overline{w\prime w\prime }}{\partial z}=0
\label{zmoment_r}
\end{equation}

where \textit{(x,y,z)} represent the space coordinate,\textit{t} represents time, \textit{$\bar u,\bar v,\bar w,\bar p$} represent the mean u, v, w component of velocity and pressure, respectively. ($\overline{u\prime u\prime},\overline{u\prime v\prime },\overline{u\prime w\prime},\overline{v\prime v\prime },\overline{v\prime w\prime },\overline{w\prime w\prime }$) indicates the Reynolds stress tensor components,and \textit{Re} denotes the Reynolds number.\\

The formulation of the loss function for this case study resembles that of the 2D unsteady laminar flow past a cylinder described in the previous sections, albeit with modifications to the PDE residual term and the data loss terms.
\begin{equation}
    L= L_{data}+L_{pde}
\label{ship_loss}
\end{equation}
$L_{pde}$ and $L_{data}$ represent the contributions from the physics loss and data loss components to the loss function computation, respectively. The components of the loss function present in  eqn.~\ref{ship_loss} are described as follows:
\begin{enumerate}
    \item $L_{pde}$: The residual loss component represents the MSE loss associated with the residual of the governing PDE, the RANS equations. The following equations \ref{s-xcomp}-\ref{s-cont} correspond to the x-momentum, y-momentum, z-momentum, and continuity residuals, respectively.
    \begin{equation}
    L_{pde_{m_x}}=\frac{1}{N_r} \sum_{k=1}^{N_r} (e_{mx}(x_{res}^k,t^k))^2
    \label{s-xcomp}
    \end{equation}
    \begin{equation}
    L_{pde_{m_y}}=\frac{1}{N_r} \sum_{k=1}^{N_r} (e_{my}(x_{res}^k,t^k))^2
    \label{s-ycomp}
    \end{equation}
    \begin{equation}
    L_{pde_{m_z}}=\frac{1}{N_r} \sum_{k=1}^{N_r} (e_{mz}(x_{res}^k,t^k))^2
    \label{s-zcomp}
    \end{equation}
    \begin{equation}
    L_{pde_{c}}=\frac{1}{N_r} \sum_{k=1}^{N_r} (e_{c}(x_{res}^k,t^k))^2
    \label{s-cont}
    \end{equation}
    \begin{equation}
        L_{pde}=L_{pde_{m_x}}+L_{pde_{m_y}}+L_{pde_{m_z}}+L_{pde_{c}}
    \label{s-phy_loss}
    \end{equation}
    
    \item $L_{data}$: Data loss represents the MSE loss estimated from the data constraints of the predictions and ground truth. 
    \begin{equation*}
     L_{u_d}=\frac{1}{N_d} \sum_{k=1}^{N_d} (\hat{u}(x_d^k,t_d^k)-u_d^k)^2
    \end{equation*}
    \begin{equation*}
     L_{v_d}=\frac{1}{N_d} \sum_{k=1}^{N_d} (\hat{v}(x_d^k,t_d^k)-v_d^k)^2
    \end{equation*}
    \begin{equation*}
     L_{w_d}=\frac{1}{N_d} \sum_{k=1}^{N_d} (\hat{w}(x_d^k,t_d^k)-w_d^k)^2
    \end{equation*}
    \begin{equation}
     L_{data}=L_{u_d}+L_{v_d}+L_{w_d}
    \label{s-data_loss}
    \end{equation}
\end{enumerate}

\subsection*{C. MODEL TRAINING}
\begin{figure*}[h!]
\hspace{-1cm} \includegraphics[width=150mm]{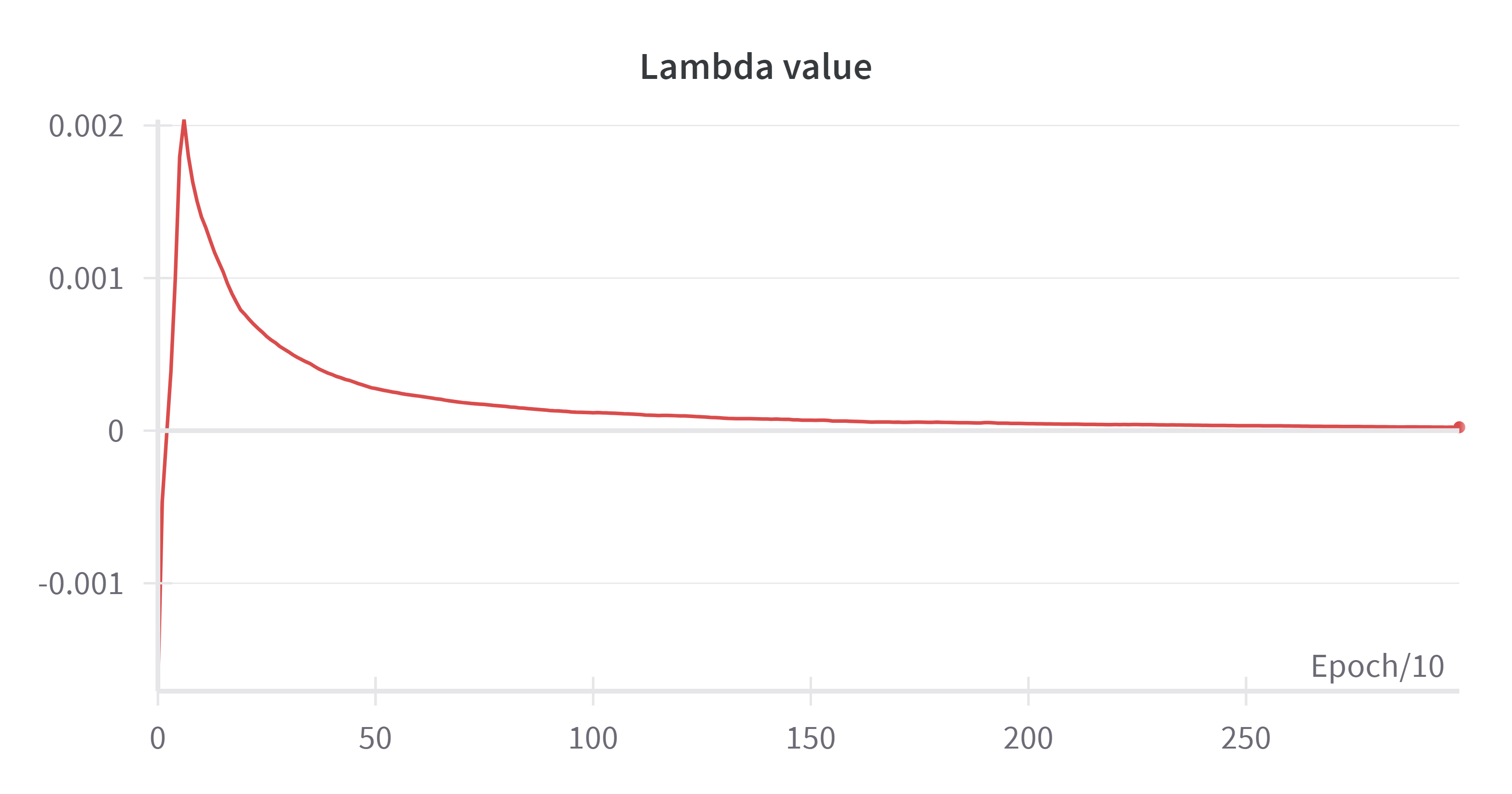}
 \caption{Convergence graph of the unknown parameter $\lambda$ as training progresses.}
 \label{fig: ship_lambda}
\end{figure*}

\begin{figure*}[h!]
\hspace{-1cm} \includegraphics[width=150mm]{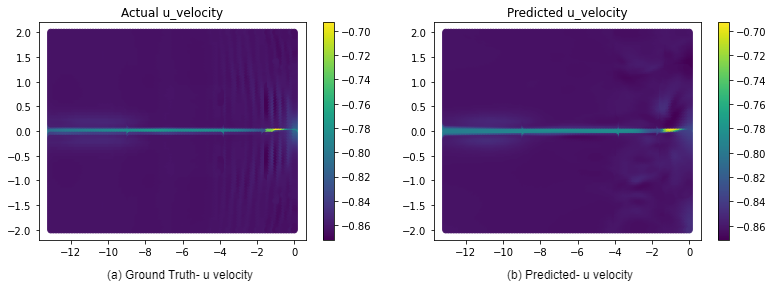}
 \caption{Comparison of ground truth and predicted u-velocity (a and b) for a test time step. } 
 \label{fig: ship_pred}
\end{figure*}

\addcontentsline{toc}{subsection}{C. MODEL TRAINING}
In the previous case study of 2D unsteady laminar flow past a circular cylinder, the Reynolds number describing the flow is a known parameter, and consequently, the governing PDE describing the physics is fully known. However, in the ULCS case study, we adopt a different approach as outlined in the data-driven discovery section of \cite{raissi2019physics}, where we do not specify the Reynolds number and instead treat it as an unknown variable to be learned from the data. Accordingly, changes are made to the governing PDE as follows:
\begin{equation}
    \frac{\partial \bar u}{\partial x}+\frac{\partial \bar v}{\partial y}+\frac{\partial \bar w}{\partial z}=0
\label{conti_s}
\end{equation}
\begin{equation}
    \frac{\partial \bar u}{\partial t}+\frac{\partial \bar p}{\partial x}+\bar u\frac{\partial \bar u}{\partial x}+\bar v\frac{\partial \bar u}{\partial y}+\bar w\frac{\partial \bar u}{\partial z}-\lambda(\frac{\partial^2 \bar u}{\partial x^2}+\frac{\partial^2 \bar u}{\partial y^2}+\frac{\partial^2 \bar u}{\partial z^2})+\frac{\partial \overline{u\prime u\prime}}{\partial x}+\frac{\partial \overline{u\prime v\prime}}{\partial y}+\frac{\partial \overline{u\prime w\prime}}{\partial z}=0
\label{xmoment_s}
\end{equation}
\begin{equation}
    \frac{\partial \bar v}{\partial t}+\frac{\partial \bar p}{\partial y}+\bar u\frac{\partial \bar v}{\partial x}+\bar v\frac{\partial \bar v}{\partial y}+\bar w\frac{\partial \bar v}{\partial z}-\lambda(\frac{\partial^2 \bar v}{\partial x^2}+\frac{\partial^2 \bar v}{\partial y^2}+\frac{\partial^2 \bar v}{\partial z^2})+\frac{\partial \overline{v\prime u\prime }}{\partial x}+\frac{\partial \overline{v\prime v\prime }}{\partial y}+\frac{\partial \overline{v\prime w\prime }}{\partial z}=0
\label{ymoment_s}
\end{equation}
\begin{equation}
    \frac{\partial \bar w}{\partial t}+\frac{\partial \bar p}{\partial z}+\bar u\frac{\partial \bar w}{\partial x}+\bar v\frac{\partial \bar w}{\partial y}+\bar w\frac{\partial \bar w}{\partial z}-\lambda(\frac{\partial^2 \bar w}{\partial x^2}+\frac{\partial^2 \bar w}{\partial y^2}+\frac{\partial^2 \bar w}{\partial z^2})+\frac{\partial \overline{w\prime u\prime }}{\partial x}+\frac{\partial \overline{w\prime v\prime }}{\partial y}+\frac{\partial \overline{w\prime w\prime }}{\partial z}=0
\label{zmoment_s}
\end{equation}
where $\lambda$ is the unknown parameter.
The physics-informed model is trained using data from sparsely sampled wake field locations. The objective is to minimize the residual of the governing PDE and comply with the data constraints. During this process, the model parameters are trained to learn the value of the unknown variable, $\lambda$. Fig. \ref{fig: ship_lambda} illustrates the convergence of the $\lambda$ value learned from the data, where the x-axis denotes the number of epochs and the y-axis denotes the lambda value. This convergence signifies that the surrogate model is able to learn the unknown coefficient from the training data. Fig. \ref{fig: ship_pred} illustrates the model's performance on a test time step where the prediction of the stream-wise velocity is compared against the ground truth available from the CFD simulation. The observations suggest that the model successfully reconstructs the complete velocity field from the sparsely sampled dataset. %Despite the surrogate model's ability to reconstruct the velocity data, the model is unable to reconstruct the pressure field.

%%%%%%%%%%%%%%%%%%%%%%%%%%%%%%%%%%%%%%%%%%%%%%%%%%%%%%%%%%%%
%%%%%%%%%%%%%%%%%%%% Ship case- URANS %%%%%%%%%%%%%%
%%%%%%%%%%%%%%%%%%%%%%%%%%%%%%%%%%%%%%%%%%%%%%%%%%%%%%%%%%%%

% The Unsteady Reynolds-Averaged Navier Stokes equation (URANS) based commercial software, Star CCM+, is employed here to simulate the 3d turbulent flow past a ULCS. A mesh size of %insert mesh size here%
% is used in the whole simulation domain of size % insert domain size here %
% The total number of cells is about % insert the number of cells %
% 30s simulations are carried out with a time step of 0.01s. From the last 10s simulations, the velocity measurements are collected.

%%%%%%%%%%%%%%%%%%%%%%%%%%%%%%%%%%%%%%%%%%%%%%%%
%%%%%%%%%%%%%%%% Conclusion %%%%%%%%%%%%%%%%%%%%
%%%%%%%%%%%%%%%%%%%%%%%%%%%%%%%%%%%%%%%%%%%%%%%%  
\section*{IV. CONCLUSION} \label{conclusion}
\addcontentsline{toc}{section}{IV. CONCLUSION}
A physics-informed surrogate model is developed for flow reconstruction from sparse datasets. In this paper, two case studies are considered to investigate the performance of the physics-informed surrogate model: (a) 2D unsteady laminar flow past a cylinder and (b) 3D unsteady turbulent flow past a ship. The performance of the surrogate model is evaluated against that of the data-driven model trained on the same dataset for reconstructing the flow field data. Hyper-parameter tuning is performed to determine the appropriate topology of the network, i.e., the number of hidden layers and the number of neurons in each hidden layer. The histograms of the back-propagated gradients are studied to understand the issue of competing optimization objectives during training. A multi-component loss weighting strategy is employed to enhance the model's performance within a constrained computational budget. Two methods of multi-component loss weighting are discussed: (a) Systematic relaxation of physics loss and (b) Adaptive weighting strategies. While the systematic relaxation strategy requires human intervention during training, the adaptive weighting strategy would limit it, thus making the surrogate model scalable. Notably, the surrogate model successfully retrieved pressure as a latent variable despite not being provided with any information about pressure during the training process. This contrasts the performance of data-driven models in the case of the 2D unsteady laminar flow past a cylinder. \\
The second case study presents an innovative approach to understanding and assessing the capabilities of physics-informed surrogate models in addressing the flow reconstruction problem for highly turbulent flow past a ship. This study employs a novel method to train the surrogate model, enabling it to learn the unknown coefficient of the governing PDE of the flow. While some initial results appear promising, further improvements in the surrogate models are necessary, particularly in terms of network architecture and training methodologies. This training methodology is particularly advantageous in scenarios where the governing PDE underlying the training data is unknown.

%%%%%%%%%%%%%%%%%%%%%%%%%%%%%%%%%%%%%%%%%%%%%%%%
%%%%%%%%%%%% Acknowledgement %%%%%%%%%%%%%%%%%%%
%%%%%%%%%%%%%%%%%%%%%%%%%%%%%%%%%%%%%%%%%%%%%%%%  
\section*{ DECLARATION OF COMPETING INTEREST}
\addcontentsline{toc}{section}{DECLARATION OF COMPETING INTEREST}
The authors have no conflicts to disclose.

\section*{ DATA AVAILABILITY STATEMENT}
\addcontentsline{toc}{section}{DATA AVAILABILITY STATEMENT}
The data that support the findings of this study are available
from the corresponding author upon reasonable request.

\section*{ AUTHOR CONTRIBUTIONS}
\addcontentsline{toc}{section}{AUTHOR CONTRIBUTIONS}
Vamsi Sai Krishna: Writing – original draft, Validation, Software, Methodology, Investigation, Formal analysis, Conceptualization. Suresh Rajendran: Writing – review \& editing, Supervision, Methodology, Investigation, Formal analysis, Conceptualization.

% \section{Acknowledgement}

%%%%%%%%%%%%%%%%%%%%%%%%%%%%%%%%%%%%%%%%%%%%%%%%
%%%%%%%%%%%%%%%%%%%%%%%%%%%%%%%%%%%%%%%%%%%%%%%%
%%%%%%%%%%%%%%%%%%%%%%%%%%%%%%%%%%%%%%%%%%%%%%%%  
\bibliographystyle{aip}
\bibliography{main}
\end{document}